\documentclass{article} 
\usepackage{colm2024_conference}

\usepackage{booktabs}
\usepackage{graphicx}
\usepackage{enumitem}
\usepackage{wrapfig}
\usepackage{algorithm}
\usepackage{algpseudocode}
\usepackage{natbib}
\usepackage{makecell}
\usepackage{booktabs}
\usepackage{bbm}
\usepackage{array}
\usepackage{amsmath} 
\usepackage{amssymb}
\usepackage{amsfonts}
\usepackage{multirow}
\usepackage{verbatim}
\usepackage{caption}
\usepackage{longtable}
\usepackage{supertabular}
\usepackage{hyperref}
\usepackage{CJKutf8}
\usepackage[utf8]{inputenc} 
\usepackage[T1]{fontenc} 
\usepackage[french,vietnamese,mongolian,greek,english]{babel}
\usepackage{pifont}
\usepackage{afterpage}
\usepackage{tablefootnote}
\usepackage{colortbl}
\usepackage{xspace}
\usepackage{textcomp}
\usepackage{makecell}
\usepackage{lscape} 
\usepackage{siunitx}
\usepackage{listings}
\usepackage{xcolor}
\usepackage{adjustbox}
\definecolor{dt}{gray}{0.7}
\definecolor{tongyi-purple}{RGB}{97,92,237}
\colorlet{tongyi-purple-alpha}{tongyi-purple!38}

\usepackage{pifont}       
\usepackage{bbding}       
\usepackage{fontawesome}

\usepackage{scrextend}

\usepackage{tgpagella}
\usepackage{latexsym}
\usepackage[T1]{fontenc}
\usepackage[utf8]{inputenc}
\usepackage{microtype}
\definecolor{mydarkblue}{rgb}{0,0.08,0.45}
\definecolor{citecolor}{HTML}{0071BC}
\usepackage{url}            
\usepackage{nicefrac}       
\usepackage{changepage}
\usepackage{xargs}          
\usepackage{wrapfig,lipsum,booktabs}
\usepackage{longtable}
\usepackage{subcaption}
\usepackage{endnotes}

\usepackage{pgfplots}
\usetikzlibrary{pgfplots.groupplots}
\pgfplotsset{compat=1.3}
\usepackage{tikz}
\usetikzlibrary{patterns}

\usepackage[most]{tcolorbox}
\usepackage{fvextra}
\usepackage{graphicx}
\usepackage[capitalize,noabbrev]{cleveref}
\crefname{section}{Section}{\S\S}
\Crefname{section}{Section}{\S\S}
\crefname{table}{Table}{Tables}
\crefname{figure}{Figure}{Figures}
\crefname{algorithm}{Algorithm}{}
\crefname{equation}{eq.}{}
\crefname{appendix}{Appendix}{}
\crefformat{section}{Section #2#1#3}
\usepackage{multicol}
\usepackage{fancyvrb}
\newsavebox{\myverbcontent}
\usepackage{titlesec}
\titleformat*{\section}{\large\bfseries}

\usepackage{nicematrix} 
\usepackage{arydshln}

\makeatletter
\DeclareRobustCommand\onedot{\futurelet\@let@token\@onedot}
\def\@onedot{\ifx\@let@token.\else.\null\fi\xspace}

\usepackage{todonotes}
\definecolor{tongyiPurple}{HTML}{643CE7}

\usepackage{tabularx}

\RecustomVerbatimEnvironment{verbatim}{Verbatim}{breaklines=true, breakanywhere=true}
\newtcolorbox{promptblock}[1]{
  breakable,
  colback=tongyiPurple!25,      
  colframe=tongyiPurple!75,    
  coltitle=black,       
  fonttitle=\bfseries, 
  title={#1},           
  arc=1mm,              
  boxrule=0.5mm,        
  left=2mm, right=2mm, top=2mm, bottom=2mm, 
  toptitle=1mm, bottomtitle=1mm 
}

\usepackage{soul}

\title{Qwen3-VL-Embedding and Qwen3-VL-Reranker: A Unified Framework for State-of-the-Art Multimodal Retrieval and Ranking}

\author{
Mingxin Li\thanks{Equal contribution.} \quad
Yanzhao Zhang\footnotemark[1] \quad
Dingkun Long\footnotemark[1] \quad
Keqin Chen \\
Sibo Song \quad
Shuai Bai \quad
Zhibo Yang \quad
Pengjun Xie \\
An Yang \quad
Dayiheng Liu \quad
Jingren Zhou \quad
Junyang Lin \\[2mm]
\textbf{Tongyi Lab, Alibaba Group}
}

\begin{document}

\maketitle

\begin{abstract} 
In this report, we introduce the Qwen3-VL-Embedding and Qwen3-VL-Reranker model series, the latest extensions of the Qwen family built on the Qwen3-VL foundation model. Together, they provide an end-to-end pipeline for high-precision multimodal search by mapping diverse modalities, including text, images, document images, and video, into a unified representation space. The Qwen3-VL-Embedding model employs a multi-stage training paradigm, progressing from large-scale contrastive pre-training to reranking model distillation, to generate semantically rich high-dimensional vectors. It supports Matryoshka Representation Learning, enabling flexible embedding dimensions, and handles inputs up to 32k tokens. Complementing this, Qwen3-VL-Reranker performs fine-grained relevance estimation for query-document pairs using a cross-encoder architecture with cross-attention mechanisms. Both model series inherit the multilingual capabilities of Qwen3-VL, supporting more than 30 languages, and are released in \textbf{2B} and \textbf{8B} parameter sizes to accommodate diverse deployment requirements. Empirical evaluations demonstrate that the Qwen3-VL-Embedding series achieves state-of-the-art results across diverse multimodal embedding evaluation benchmarks. Specifically, Qwen3-VL-Embedding-8B attains an overall score of \textbf{77.8} on MMEB-V2, ranking first among all models (as of January 8, 2025). This report presents the architecture, training methodology, and practical capabilities of the series, demonstrating their effectiveness on various multimodal retrieval tasks, including image-text retrieval, visual question answering, and video-text matching.
\end{abstract} 

\begin{figure}[ht]
    \centering
    \includegraphics[width= 1\linewidth]{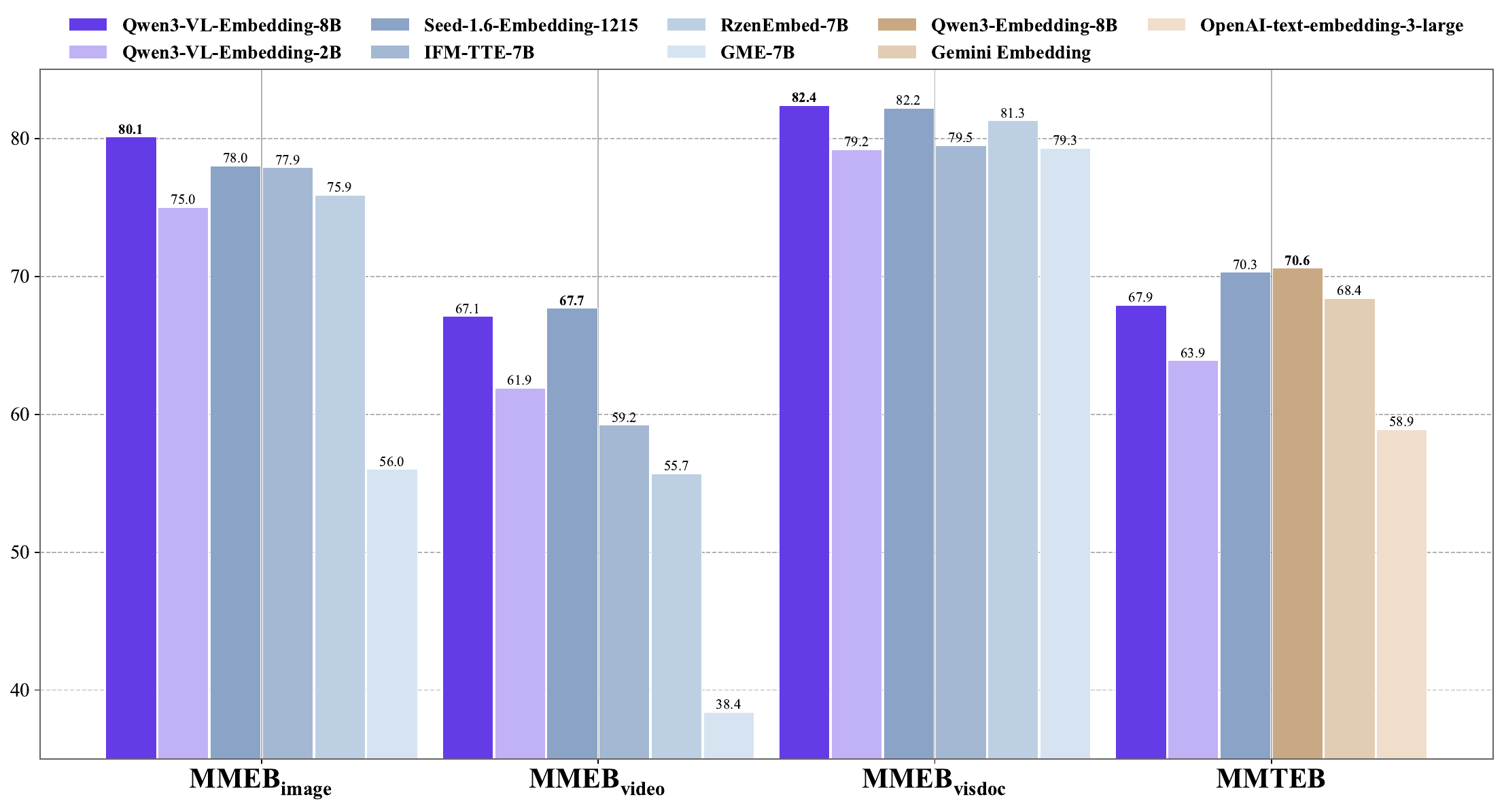}
\end{figure}

\begin{figure}[ht]
    \centering
    \includegraphics[width= 1\linewidth]{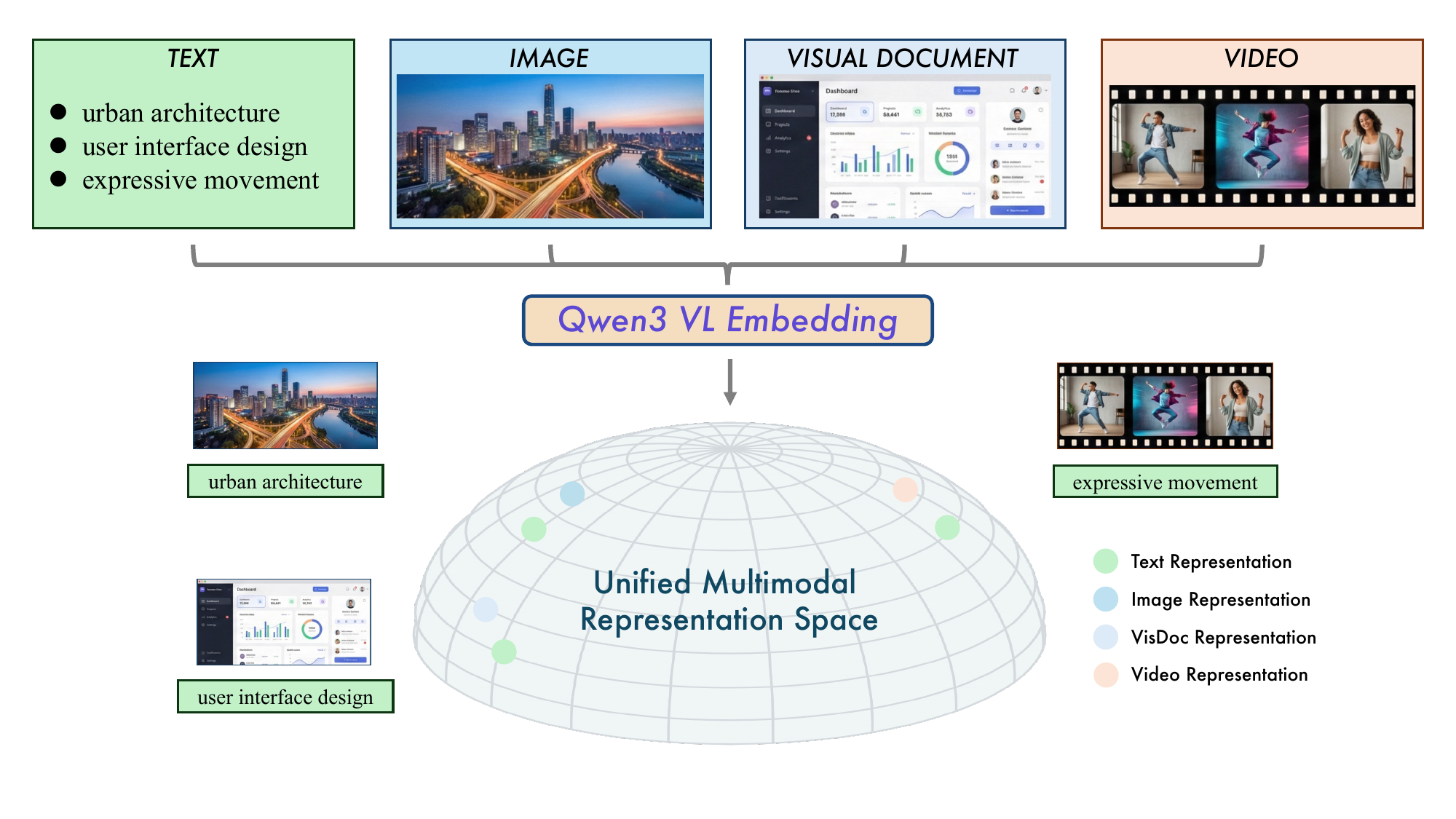}
    \caption{Illustration of the Unified Multimodal Representation Space. Qwen3-VL-Embedding model series represent multi-source data (Text, Image, Visual Document, and Video) into a common manifold. By aligning semantic concepts across modalities (e.g., the text "urban architecture" and its corresponding image), the model achieves a holistic understanding of complex visual and textual information.}
\end{figure}

\section{Introduction}
The exponential growth of multimodal content on the internet has fundamentally transformed how information is created, shared, and consumed. Modern digital ecosystems are increasingly populated with diverse data modalities, including natural images, text documents, infographics, screenshots, and videos. This proliferation necessitates advanced retrieval systems capable of understanding and matching semantic concepts across different modalities, moving beyond traditional text-only search paradigms. Multimodal search, which aims to retrieve relevant content regardless of the query or document modality, has emerged as a critical capability for applications ranging from e-commerce product discovery to scientific literature exploration and social media navigation~\citep{faysse2025colpali,fu2025moon}.

Within contemporary multimodal retrieval architectures, embedding and reranking models constitute the two most critical modules. The field of multimodal representation learning has witnessed significant progress over the past decade~\citep{manzoor2023multimodality,mei2025survey}. Among these pioneering works, CLIP (Contrastive Language-Image Pre-training)~\citep{radford2021learning} has been particularly influential by demonstrating that large-scale contrastive learning on image-text pairs can produce powerful aligned representations. Its success has cemented the importance of learning shared embedding spaces where semantically similar content is positioned proximate in the representation space regardless of its modality. 

As the development of foundation models accelerates, multimodal pre-trained vision-language models (VLMs) such as Qwen-VL~\citep{wang2024qwen2,Qwen3-VL} and GPT-4o~\citep{hurst2024gpt} have achieved unprecedented success in multimodal comprehension. Building on these breakthroughs, the multimodal retrieval community has increasingly explored training unified multimodal embedding models based on VLMs. Notable efforts in this space include E5-V~\citep{jiang2024e5}, GME~\citep{zhang2025bridging}, BGE-VL~\citep{zhou2025megapairs}, and VLM2Vec~\citep{meng2025vlm2vec,jiang2025vlm2vec}. Training unified multimodal representations based on VLMs offers several compelling advantages. First, VLMs possess inherent cross-modal alignment through their pre-training on large-scale image-text datasets. Second, they leverage sophisticated attention mechanisms to capture fine-grained interactions between visual and textual elements. Third, they provide a natural pathway to handling complex multimodal documents such as infographics and presentation slides where visual and textual information are deeply intertwined. Furthermore, VLM-based approaches can inherit the extensive multilingual and multi-domain knowledge encoded in foundation models, enabling more robust generalization across diverse retrieval scenarios.

In this work, we introduce the Qwen3-VL-Embedding and Qwen3-VL-Reranker model series, which are specifically designed for multimodal retrieval applications. Built upon the powerful Qwen3-VL~\citep{Qwen3-VL} foundation model, these models bring together advanced vision-language understanding capabilities with specialized training methodologies tailored for retrieval tasks. The Qwen3-VL-Embedding series employs a sophisticated multi-stage training paradigm that progresses from contrastive pre-training on large-scale multimodal data to knowledge distillation from ranking models, ultimately producing semantically rich embeddings that capture nuanced relationships across modalities. These models support Matryoshka Representation Learning~\citep{kusupati2022matryoshka}, allowing users to flexibly select embedding dimensions according to their storage and computational constraints without retraining. Additionally, we incorporate quantization-aware training strategies during the training process to ensure that the generated embeddings maintain robust performance after quantization. This capability significantly improves the storage efficiency and computational friendliness of downstream tasks. The models can process inputs containing up to 32,768 tokens, enabling comprehensive understanding of long documents and videos. Complementing the embedding models, the Qwen3-VL-Reranker series adopts a cross-encoder architecture that performs deep cross-attention between query and document representations, providing precise relevance scores for candidate retrieval results. Both model series inherit the impressive multilingual capabilities of the Qwen3-VL foundation model, supporting more than 30 languages with high proficiency, and are released in two sizes (2B, and 8B parameters) to accommodate diverse application scenarios.

We evaluate the Qwen3-VL-Embedding and Qwen3-VL-ReRanker model series across a comprehensive set of benchmarks spanning multiple tasks and domains. Experimental results demonstrate that our embedding and reranking models achieve state-of-the-art performance across multiple types of downstream tasks. For example, the flagship model Qwen3-VL-Embedding-8B attains a score of 77.8 on the MMEB-V2 benchmark~\citep{meng2025vlm2vec}, as evaluated in January 2026, surpassing all models currently on the leaderboard\footnote{\url{https://huggingface.co/spaces/TIGER-Lab/MMEB-Leaderboard}}, including both open-source models and closed-source API services. Beyond multimodal evaluation, in pure text evaluation, the Qwen3-VL-Embedding-8B model achieves a mean task score of 67.9 on the MTEB Multilingual benchmark~\citep{enevoldsen2025mmteb}, demonstrating highly competitive performance. Moreover, our reranking model delivers competitive results across a range of retrieval tasks. The Qwen3-VL-Reranker-2B model exceeds previously top-performing models in numerous retrieval tasks, while the larger Qwen3-VL-Reranker-8B model demonstrates even superior performance, improving ranking results by 4.1 points over the 2B model across multiple tasks. Furthermore, we include a constructive ablation study to elucidate the key factors contributing to the superior performance of the Qwen3-VL-Embedding series, providing insights into its effectiveness.

In the following sections, we present the architectural design of our model, elaborate on the training procedures, report comprehensive experimental results for both the embedding and reranking components, and conclude this technical report by synthesizing key findings and discussing promising avenues for future investigation.

\begin{figure}[t]
\centering
\includegraphics[width= 1\linewidth]{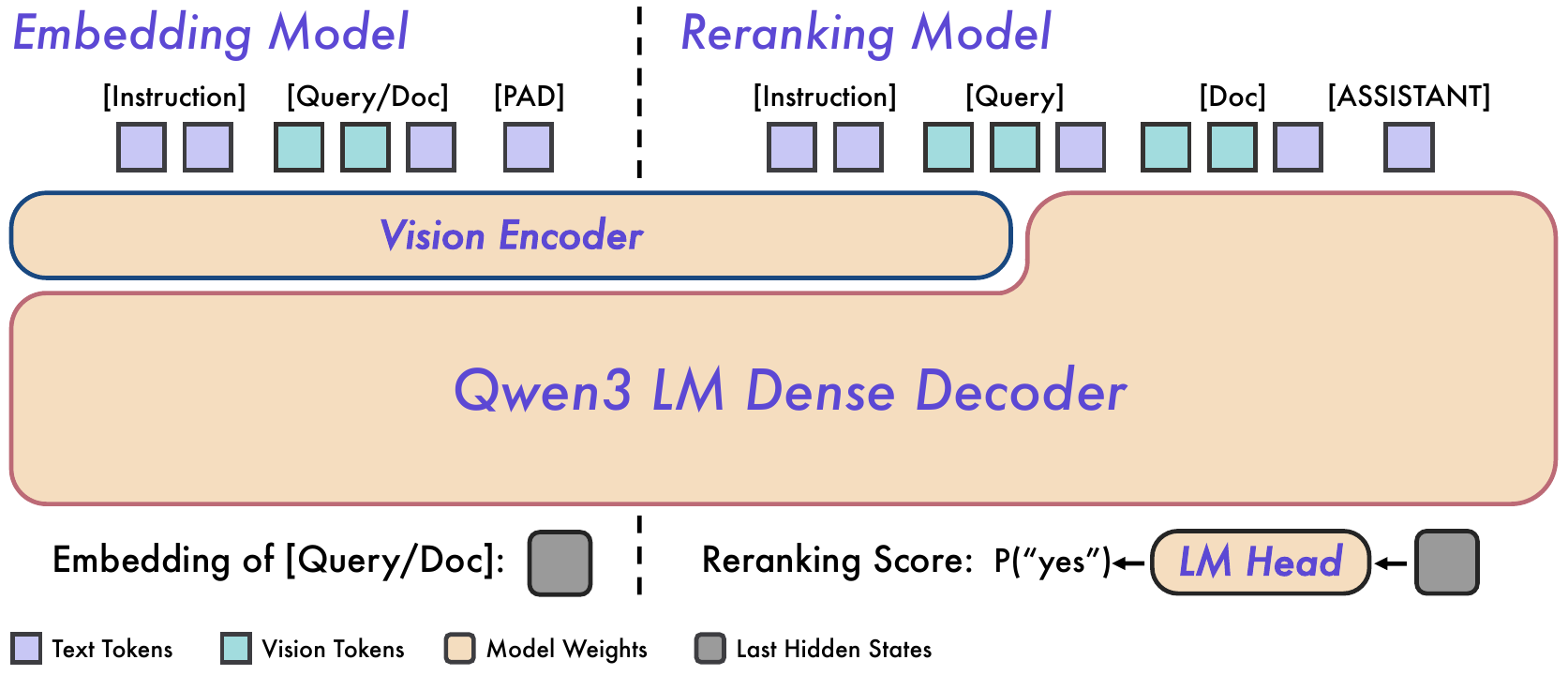}
   \caption{
   Overview of the Qwen3-VL-Embedding and Qwen3-VL-Reranker architecture.
}
\label{fig:arc}
\end{figure}

\section{Model}

\begin{table}[!t]
	\centering
	\renewcommand{\arraystretch}{1.2}
	\small
	\caption{Model specifications for the Qwen3-VL-Embedding and Qwen3-VL-Reranker. ``Quantization Support'' indicates the supported quantization formats for the embeddings. ``MRL support'' denotes whether the embedding model allows user-specified embedding dimensionalities. ``Instruction-aware'' indicates whether the models support task-specific customization of the input instruction.
}
\label{tab:model_specs}
\begin{tabular}{cccccccc}
\hline
Model Type & Size & Layers 
& \begin{tabular}{@{}c@{}}Sequence\\Length\end{tabular} 
& \begin{tabular}{@{}c@{}}Embedding\\Dimension\end{tabular} 
& \begin{tabular}{@{}c@{}}Quantization\\Support\end{tabular} 
& \begin{tabular}{@{}c@{}}MRL\\Support\end{tabular} 
& \begin{tabular}{@{}c@{}}Instruction\\Aware\end{tabular} \\
\hline
\multirow{2}{*}{Qwen3-VL-Embedding} & 2B & 28 & 32K & 2048 & Yes & Yes & Yes \\
& 8B & 36 & 32K & 4096 & Yes & Yes & Yes \\
\hline
\multirow{2}{*}{Qwen3-VL-Reranker} & 2B & 28 & 32K & - & - & - & Yes \\
& 8B & 36 & 32K & - & - & - & Yes \\
\hline
\end{tabular}
\end{table}

Qwen3-VL-Embedding and Qwen3-VL-Reranker models are designed to make task-aware relevance judgments for multimodal instances. As shown in Figure~\ref{fig:arc}, the embedding model follows a bi-encoder architecture to produce dense vector representations of instances and uses cosine similarity as the relevance measure. In contrast, the reranking model adopts a cross-encoder architecture to provide more fine-grained relevance estimates for each query–document pair.

\paragraph{Model Architecture} Both the embedding and reranking models are built on the Qwen3-VL backbone, using causal attention. After being trained on a large-scale collection of multimodal, multi-task relevance data, they retain the backbone's world knowledge, multimodal perception, and instruction-following capabilities, while additionally gaining the ability to estimate relevance. We train two model sizes—2B and 8B—and summarize their specifications in Table~\ref{tab:model_specs}.

\paragraph{Embedding Method} The embedding model extracts task-aware dense vectors for multimodal inputs. The input format follows the Qwen3-VL context structure, where the instruction is passed as a system message, with the default instruction being ``Represent the user's input.'' The multimodal instance to be represented is passed as a user message, and it can be in the form of text, images, videos, or any combination of these modalities. Finally, a ``PAD'' (\verb!<|endoftext|>!) token is appended to the input, and the last hidden state corresponding to this token is used as the dense vector representation of the instance.

\begin{promptblock}{Input Template for Embedding}
\begin{verbatim}
<|im_start|>system
{Instruction}
<|im_end|>
<|im_start|>user
{Instance}
<|im_end|>
<|im_start|>assistant
<|endoftext|>
\end{verbatim}
\end{promptblock}

\paragraph{Reranking Method} The reranking model adopts a pointwise ranking approach, which evaluates the relevance between a pair of multimodal instances according to the relevance definition provided in the instruction. The input format follows the Qwen3-VL context structure, where both the relevance-defining instruction and the pair of multimodal instances to be evaluated are passed as user messages. These multimodal inputs can be text, images, videos, or any combination of these modalities. Finally, the relevance estimation for the pair is obtained by calculating the model's probability of predicting ``yes'' or ``no'' as the next output token.

\begin{promptblock}{Input Template for Reranking}
\begin{verbatim}
<|im_start|>system
Judge whether the Document meets the requirements based on the Query and the Instruct provided. Note that the answer can only be "yes" or "no".
<|im_end|>
<|im_start|>user
<Instruct>: {Instruction} 
<Query>: {Query} 
<Document>: {Document}
<|im_end|>
<|im_start|>assistant
\end{verbatim}
\end{promptblock}

\section{Data}

To endow the model with universal representation capabilities across diverse modalities, tasks, and domains, we curated a massive-scale dataset. The distribution of different categories within the dataset is illustrated in Figure~\ref{fig:data_categories}.
However, both publicly available and proprietary in-house data exhibit significant imbalances and, in specific scenarios, notable scarcity across these dimensions. To address these challenges, we leverage data synthesis to construct a balanced training corpus that ensures robust coverage across all modalities, tasks, and domains.

\subsection{Dataset Format}

The complete dataset comprises multiple sub-datasets, denoted as $\mathcal{D} = \{D_i\}_{i=1}^M$. Each sub-dataset $D_i$ is defined by a quadruple $D_i = (I_i, Q_i, C_i, R_i)$, structured as follows:

\begin{itemize}
    \item \textbf{Instruction} ($I_i$): A textual description defining the specific relevance criteria and task objectives for the sub-dataset.
    \item \textbf{Queries} ($Q_i$): A collection of $N_q$ query objects, $Q_i = \{q_{j}\}_{j=1}^{N_q}$. Each $q_{j}$ can consist of text, images, videos, or any multimodal combination thereof.
    \item \textbf{Corpus} ($C_i$): A repository of $N_d$ document objects, $C_i = \{d_{j}\}_{j=1}^{N_d}$. Similar to queries, each $d_{j}$ may be a single modality or a multimodal composite of text, images, and videos.
    \item \textbf{Relevance Labels} ($R_i$): This component identifies the relationships between queries and documents, denoted as $R_i = \{(q_{j}, \{d_{j,k}^+\}_{k=1}^{n^+}, \{d_{j,k}^-\}_{k=1}^{n^-})\}_{j=1}^{N_q}$. For each query $q_{j}$, $\{d_{j,k}^+\}_{k=1}^{n^+} \subset C_i$ represents the set of relevant documents (positive documents), while $\{d_{j,k}^-\}_{k=1}^{n^-} \subset C_i$ represents the set of irrelevant documents (negative documents).
\end{itemize}

Representative dataset examples are presented in Appendix~\ref{sec:appendix_dataset_examples}.

\subsection{Data Synthesis} 

\begin{figure}[H]
\centering
\begin{minipage}{0.33\textwidth}
    \centering
    \includegraphics[height=5.4cm]{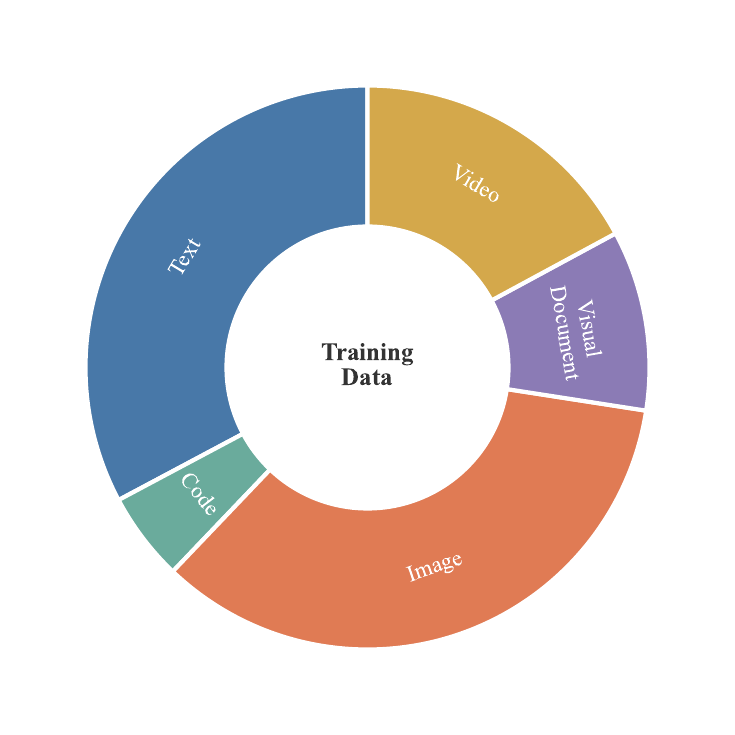}
    \caption{Distribution of different categories in the training data.}
    \label{fig:data_categories}
\end{minipage}
\hfill
\begin{minipage}{0.66\textwidth}
    \centering
    \includegraphics[height=4.9cm]{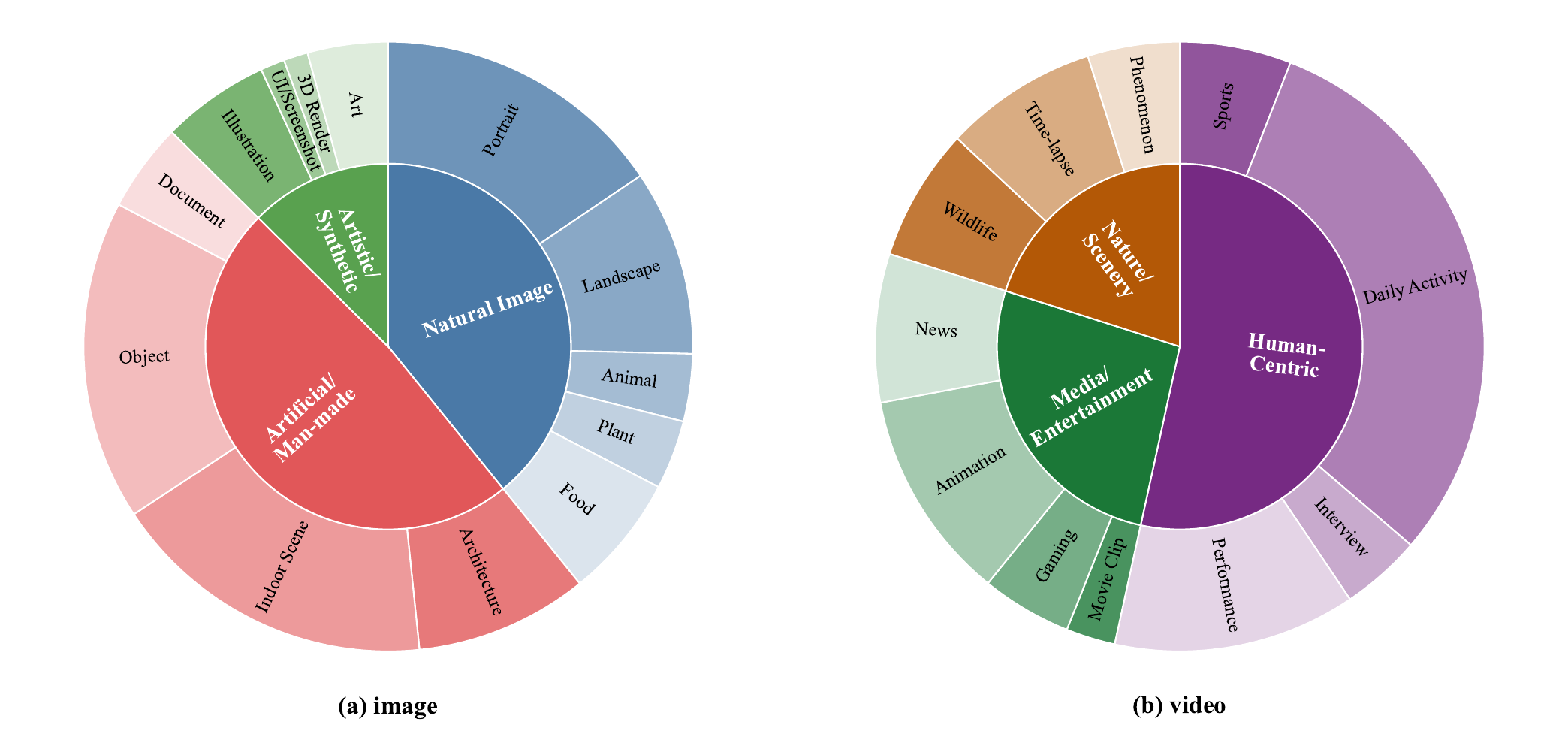}
    \caption{Data distribution of the seed pool for data synthesis.}
    \label{fig:data_distribution}
\end{minipage}
\end{figure}

We employ data synthesis to construct various sub-datasets $D_i$. Specifically, we extend the methodology introduced in Qwen3 Embedding~\citep{qwen3embedding} to multimodal scenarios. 

\paragraph{Seed Pool Construction} Since the diversity of synthesized data depends on the underlying seed pool, we first aggregate an extensive collection of high-quality and diverse raw image and video datasets. To establish a high-quality foundation, we first apply coarse-grained quality filtering to prune assets with low resolutions or irregular aspect ratios. This is followed by structural refinement, specifically employing scene cut detection and removing static or corrupted segments, to preserve the integrity of temporal dynamics in video data. Subsequently, we leverage Qwen3-VL-32B~\citep{Qwen3-VL} to generate fine-grained categorical labels for the remaining assets. To ensure cross-modal alignment, we implement a rigorous filtering mechanism that excludes samples with low-confidence annotations or poor visual-text correspondence, as measured by similarity scores from the GME~\citep{zhang2025bridging} embedding model. Finally, we perform category-wise rebalancing on the refined dataset to construct the final seed pool. The resulting category distribution is illustrated in Figure \ref{fig:data_distribution}.

Based on the seed pool, we leverage Qwen3-VL-32B~\citep{Qwen3-VL} to perform multimodal and multi-task annotation. 

\paragraph{Image Tasks Annotation} We synthesize image datasets across three primary task paradigms:
\begin{enumerate}
    \item \textbf{Image Classification}: The query $q$ comprises an image and a classification instruction, while the document $d$ is the specific category label. We synthesize datasets for a wide range of classification tasks, including object recognition, scene parsing, landmark identification, and action recognition. For each sample, the model designates a specific task type and annotates the image with its ground-truth category along with a semantically confusing negative label. 
    \item \textbf{Image Question Answering}: The query $q$ consists of an image and a grounded question, and the document $d$ is the corresponding answer. We generate diverse QA pairs covering factoid identification, visual reasoning, OCR-based data extraction, and domain-specific knowledge inquiry. Following a prescribed task orientation, the model formulates a question based on the visual content, providing a ground-truth response and a plausible but deceptive distractor. 
    \item \textbf{Image Retrieval}: The query $q$ is a search text, and the document $d$ is the candidate image. We synthesize retrieval queries across a hierarchy of semantic depths, spanning direct visual descriptions, abstract narrative scenarios, compositional logical constraints, and knowledge-centric textual localization. The model assigns a specific retrieval intent and generates a corresponding search query that captures either the salient visual features or the embedded textual logic within the image. 
\end{enumerate}

\paragraph{Video Tasks Annotation} We synthesize video datasets across four primary task paradigms:
\begin{enumerate}
    \item \textbf{Video Classification}: The query $q$ combines a video with a classification task, and the document $d$ is the resulting category. We synthesize datasets for diverse classification tasks, including activity recognition, scene parsing, event categorization, and sentiment/intent analysis. For each sample, the model identifies its category and generates a semantically related negative label. 
    \item \textbf{Video Question Answering}: The query $q$ includes a video and a question, while the document $d$ is the answer. We generate diverse QA pairs spanning factual identification, temporal grounding, thematic reasoning, and cinematic analysis. Guided by a specified task type, the model formulates a question and provides a correct response and a deceptive distractor.
    \item \textbf{Video Retrieval}: The query $q$ is a textual description, and the document $d$ is the video. We synthesize retrieval queries across a spectrum of semantic granularities, ranging from entity and action-centric searches to temporal-event descriptions, thematic/emotional discovery, and instructional tutorial localization. The model produces a search query that captures the primary events and thematic content of the video.
    \item \textbf{Moment Retrieval}:  The query $q$ is a textual query (optionally including a keyframe), and the document $d$ is a specific video segment. The moment retrieval task aims at fine-grained temporal grounding. The model identifies a specific target—such as an action, object, or character—and localizes a relevant temporal segment. Simultaneously, it identifies an irrelevant segment with a clear temporal gap to serve as a negative contrast.
\end{enumerate}
Prior to synthesizing task-specific annotations, we require the model to generate a descriptive caption for each image or video to provide necessary context. This two-step approach ensures higher quality and consistency in the subsequent annotation generation. Selected prompt examples for the synthesis of specific tasks are provided in Appendix~\ref{sec:data_synthesis_prompts}.

\subsection{Positive Refinement and Hard Negative Mining}
\label{sec:data_mining}

Hard negative samples play a crucial role in contrastive representation learning~\citep{robinson2021contrastive}. To enhance the quality of positive pairs and identify effective hard negatives, we implement an automated two-stage mining pipeline: Recall and Relevance Filtering.

\paragraph{Recall} For each sub-dataset $D_i$, we use an embedding model to extract representations for all queries $q_j \in Q_i$ and documents $d_k \in C_i$. For each query $q_j$, we retrieve the top-$K$ most relevant candidates $\{d_k\}_{k=1}^K$ based on cosine similarity, denoted as relevance scores $S = \{s_{j,k}\}_{k=1}^K$.

\paragraph{Relevance Filtering} Finally, we refine the relevance labels $R_i$ based on the relevance scores $S$ to eliminate noise:
\begin{itemize}
    \item \textbf{Positive Refinement}: We retain $q_j$ only if at least one positive document $d^+ \in \{d_k\}_{k=1}^{K}$ achieves a score $s > t^+$, where $t^+$ is a hyperparameter acting as the score threshold. If no such candidate exists, the query $q_j$ is discarded.
    \item \textbf{Hard Negative Selection}: For a valid query $q_j$, we compute the average score of its refined positive samples, $\bar{s}^+$. Any non-positive document $d \in \{d_k\}_{k=1}^{K}$ is selected as a hard negative only if its score satisfies $s < \bar{s}^+ + \delta^-$, where $\delta^-$ is a small safety margin to prevent the inclusion of ``false negatives''.
\end{itemize}

\section{Training Strategy}

\begin{figure}[t]
\centering
\includegraphics[width= 1\linewidth]{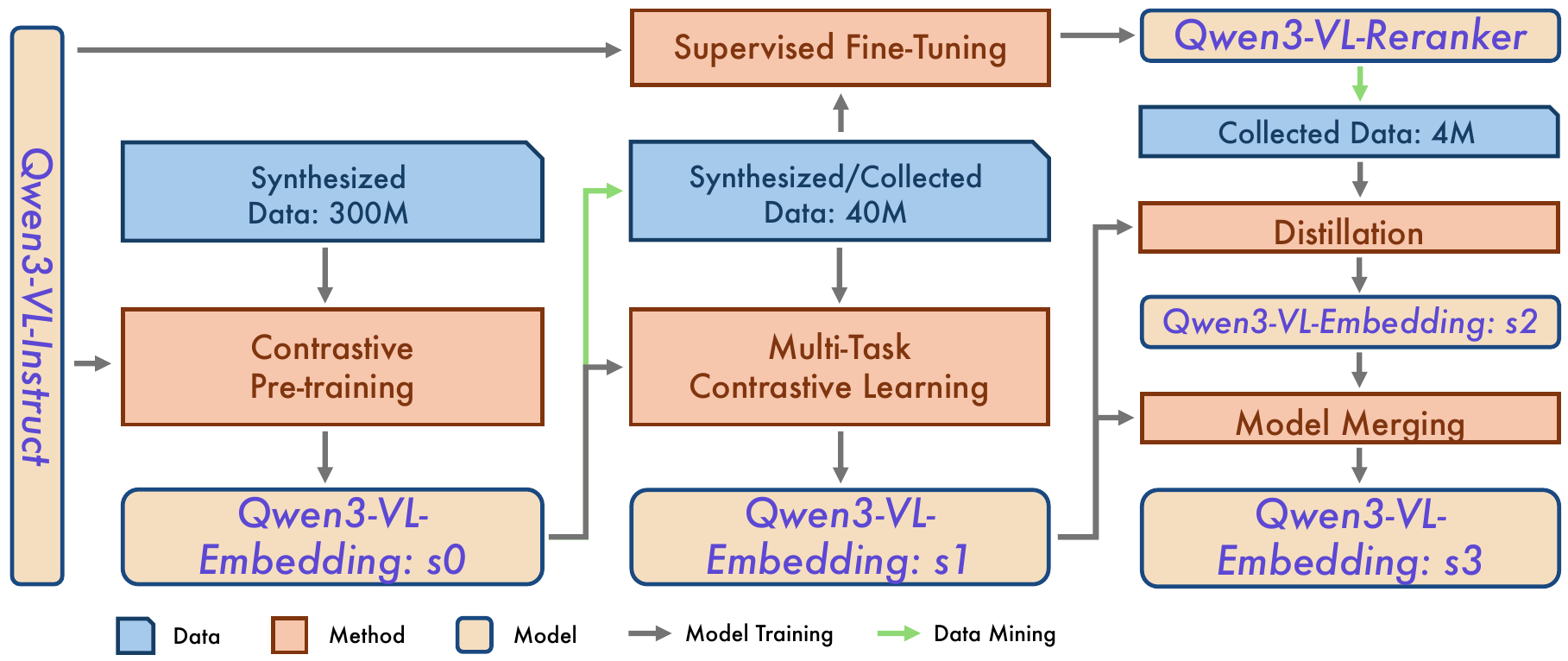}
   \caption{
   The multi-stage training pipeline of Qwen3-VL-Embedding and Qwen3-VL-Reranker.
}
\label{fig:pipline}
\end{figure}

To train our Qwen3-VL-Embedding and Qwen3-VL-Reranker, we employ a multi-stage training pipeline, as shown in Figure~\ref{fig:pipline}. This approach is designed to mitigate the data imbalance between abundant weakly-supervised data and scarce high-quality samples~\citep{wang2022text, li2023towards, chen-etal-2024-m3, qwen3embedding}. 
The model is first pre-trained on vast amounts of weakly supervised, noisy data to establish a baseline for relevance understanding and to boost generalization. We then perform fine-tuning on high-quality, task-specific datasets to steer the model toward more precise relevance scoring and fine-grained interaction. 
In addition to the aforementioned reasons, another objective of the multi-stage training strategy is to bootstrap both data quality and model performance. As the training progresses through successive stages, the model's capabilities are continuously enhanced. This improvement, in turn, facilitates more effective data mining, thereby refining the quality of the training data. This iterative cycle ultimately leads to a substantial boost in the model's overall performance.

\subsection{Multi-stage Training}

We implement a three-stage training strategy as follows:

\paragraph{Stage 1: Contrastive Pre-training}\label{sec:stage1} 
In the pre-training stage, we first perform contrastive learning on the embedding model using large-scale, multimodal, and multi-task synthetic data. 
The synthetic data utilized in this stage is mined using the methodology described in Section~\ref{sec:data_mining}, utilizing an existing open-source model~\citep{zhang2025bridging} as the embedding model.

The optimization objective employed during training is defined in Equation~\ref{loss:retrieval}. Upon completion of this stage, we obtain the initial model version, Qwen3-VL-Embedding: s0.

\paragraph{Stage 2: Multi-Task Contrastive Learning and Supervised Fine-Tuning}\label{sec:stage2}
In this stage, we primarily utilize a combination of curated public datasets and proprietary in-house data, augmented with sampled synthetic data to address the task imbalance inherent in existing datasets. Benefiting from the improved multi-task performance of Qwen3-VL-Embedding: s0, we employ this model to perform data mining, thereby ensuring high data quality across various tasks. We then train our embedding model using multi-task contrastive learning, implementing tailored contrastive objectives for different task types (see Section~\ref{sec:losses_embedding} for details). This results in Qwen3-VL-Embedding: s1.

Simultaneously, we train a new reranking model, Qwen3-VL-Reranker, by training on the retrieval-specific subset of the newly mined data, using Equation~\ref{loss:reranking} as the optimization objective. This subset encompasses diverse tasks, including image retrieval, video retrieval, moment retrieval, and visual document retrieval. The resulting model demonstrates superior performance across these retrieval-centric tasks.

\paragraph{Stage 3: Distillation and Model Merging}\label{sec:stage3}
In this final stage, we further enhance the embedding model by distilling the relevance discrimination expertise from the previously trained Qwen3-VL-Reranker. To achieve this, we curate a compact sub-dataset from both public and proprietary sources, ensuring a balanced distribution across multiple retrieval categories. We then employ Qwen3-VL-Reranker to generate fine-grained relevance scores for this subset, which serves as the supervision signal for training the embedding model under the objective defined in Equation~\ref{loss:distillation}. This distillation process yields Qwen3-VL-Embedding: s2.

While Qwen3-VL-Embedding: s2 exhibits significant gains in retrieval-centric tasks, it suffers a slight performance degradation in classification and QA tasks. To address this, we merge Qwen3-VL-Embedding: s2 with Qwen3-VL-Embedding: s1 using the methodology proposed by \cite{li2024improving}. This process results in our final model, Qwen3-VL-Embedding: s3, which achieves optimal and balanced performance across all evaluated tasks.

\subsection{Implementation}

We employ Low-Rank Adaptation (LoRA)~\citep{DBLP:conf/iclr/HuSWALWWC22} for model training, with the model parameters initialized from Qwen3-VL-Instruct. This approach offers several key advantages: 1) reduced memory footprint, allowing for larger effective batch sizes; 2) enhanced generalization performance; and 3) significantly more efficient hyperparameter search for model merging~\citep{li2024improving}. 
Additionally, we adopt dynamic resolution and frame rates. For the image modality, we preserve the original aspect ratio while capping the maximum token consumption at 1,280 (approximately $1.3 \times 10^6$ pixels). For video, we first sample at 1 FPS with a maximum of 64 frames. For each frame, the aspect ratio is maintained, and the total token budget for all frames is constrained to 4,500 (approximately $9.2 \times 10^6$ pixels).

\section{Training Objective}

This section outlines the training objectives for the Qwen3-VL-Embedding and Qwen3-VL-Reranker.  
For Qwen3-VL Embedding model, We extend the loss function from the Qwen3 Embedding model~\citep{qwen3embedding} to handle a wider variety of data types. 
We also integrate two key techniques: Matryoshka Representation Learning (MRL)~\citep{kusupati2022matryoshka} to produce variable-dimension embeddings, and Quantization-Aware Training (QAT)~\citep{esser2020learned} to support multiple numerical precisions. Together, these methods reduce storage and compute costs, improving inference efficiency. The Qwen3-VL-Reranker adopts the same objective function as Qwen3 Reranker~\citep{qwen3embedding}. The specific loss functions for each model are detailed below.

\subsection{Loss Functions for the Embedding Model}
\label{sec:losses_embedding}

The training of Qwen3-VL-Embedding involves diverse data types across multiple stages. To accommodate this, we employ distinct loss function tailored to the specific characteristics of each data category.

\paragraph{Loss for Retrieval Data} This category includes data from various multimodal and cross-modal retrieval tasks, such as Text-to-Text (T2T), Text-to-Image (T2I), and Image+Text-to-Image+Text (IT2IT) retrieval.
In Stage 1, we use the same InfoNCE loss~\citep{oord2018representation} formulation as in the Qwen3-Embedding:
\begin{equation}
\label{loss:retrieval}
    \mathcal{L}_\textrm{retrieval} = - \frac{1}{N} \sum_i^N \log\frac{e^{(s(q_i, d_i^+)/\tau)}}{Z_i},
\end{equation}
where $s(\cdot, \cdot)$ is a similarity function (we use cosine similarity), $\tau$ is a temperature parameter, and $Z_i$ aggregates scores from the positive pair and various types of negative pairs:
\begin{equation*}
    Z_i = e^{(s(q_i, d_i^+) / \tau)}  + \sum_k^K m_{ik}e^{(s(q_i, d_{i,k}^-)/\tau)} + \sum_{j\neq i} m_{ij}e^{(s(q_i, q_j) / \tau)} + \sum_{j\neq i} m_{ij}e^{(s(d_i^+, d_j) / \tau)} + \sum_{j\neq i} m_{ij}e^{(s(q_i, d_j) / \tau)} 
\end{equation*}
corresponding to similarities with (1) the positive document $d_i^+$, (2) $K$ hard negatives $\{d_{i,k}^-\}_{k=1}^{K}$, (3) other in-batch queries $\{q_j\}_{j\neq i}$, (4) other in-batch documents $\{d_j\}_{j\neq i}$ contrasted with $d_i^+$, and (5) other in-batch documents $\{d_j\}_{j\neq i}$ contrasted with $q_i$. 
$m_{ij}$ is a masking factor to mitigate the impact of false negatives:
\begin{equation*}
m_{ij}=
\begin{cases}
0, & \text{if } s_{ij} > s(q_i,d_i^+) + 0.1 \text{ or } d_j = d_i^+,\\
1, & \text{otherwise},
\end{cases}
\end{equation*}
where $s_{ij}$ denotes the corresponding similarity score (e.g., $s(q_i,d_j)$ or $s(q_i,q_j)$).

In Stage 2, we further modify the objective by removing the query--query and document--document terms from $Z_i$. Empirically, this adjustment yields better performance on high-quality multimodal retrieval data.

\paragraph{Loss for Classification Data} For text or image classification tasks, we likewise formulate training as contrastive learning. Specifically, the instance to be classified is treated as a query $q$, and its class label is treated as the corresponding document $d^+$. In contrast to retrieval, negative samples are restricted to explicitly incorrect labels for the same query, while other labels in the batch are ignored to avoid introducing false negatives.

\paragraph{Semantic Textual Similarity (STS) Data} STS datasets are symmetric and thus do not admit a natural query--document asymmetry. Moreover, supervision is typically provided as real-valued similarity scores. To exploit this fine-grained signal, we optimize the model with the CoSent loss~\citep{10.1109/TASLP.2024.3402087}, which encourages cosine similarities between paired embeddings to preserve the ordering induced by ground-truth similarity scores:
\begin{equation}
\label{cosent}
\mathcal{L}_{\mathrm{sts}}
=
\log\!\left(
1 + \sum_{\hat{s}(q_i,d_j) > \hat{s}(q_m,d_n)}
\exp\!\left(
\frac{\cos(q_m,d_n) - \cos(q_i,d_j)}{\tau}
\right)
\right),
\end{equation}

where $\hat{s}(q_i, d_j)$ denotes the ground-truth score for the pair $(q_i, d_j)$.

\paragraph{Distillation Data}
In the final training stage, we further improve the embedding model via knowledge distillation. We sample a high-quality subset from the union of all training data and use a strong reranker to provide supervision. Concretely, for each query $q$, we pre-compute (offline) reranker relevance logits for its positive document and $k$ negatives. During training, we compute embedding-based scores online using cosine similarity and minimize a distribution-matching objective (cross-entropy) to align the embedding model's score distribution with that of the reranker:
\begin{equation}\label{loss:distillation}
\mathcal{L}_{\text{distill}}
=
-\sum_{i=1}^{k+1} P_{\text{reranker}}(d_i \mid q)\,
\log P_{\text{embedding}}(d_i \mid q),
\end{equation}
where $P(d_i \mid q)$ is the softmax distribution over the $(k{+}1)$ candidate documents (one positive and $k$ negatives) for query $q$.

\subsubsection{Additional Techniques for Efficient Inference}
In practical retrieval systems, index construction requires storing a large number of embeddings offline. To reduce storage overhead and improve retrieval efficiency, we incorporate the following auxiliary training objectives.

\paragraph{Matryoshka Representation Learning (MRL)}
When optimizing the objectives described above, we compute each loss not only on the full-dimensional embedding, but also on truncated lower-dimensional prefixes of the same representation~\citep{kusupati2022matryoshka}. Empirically, training over a sufficiently dense set of MRL dimensions yields strong generalization, enabling competitive performance at intermediate dimensions that are not explicitly included during training.

\paragraph{Quantization-Aware Training (QAT)}
Storing embeddings with lower numerical precision (int8 or binary) can further reduce both storage and compute overhead. To preserve embedding quality under low-precision representations, we adopt a quantization-aware training (QAT) strategy. Concretely, during training we compute the optimization objective using both full-precision embeddings and their low-precision (quantized) counterparts, so that the model learns to produce embeddings that are robust to quantization. This allows the learned representations to better adapt to low-bit embedding formats, mitigating the performance degradation that may otherwise occur at deployment time. 
We instantiate QAT with Learned Step Size Quantization (LSQ)~\citep{esser2020learned}. LSQ treats the quantization scale (step size) as a learnable parameter and optimizes it jointly with the model weights via backpropagation. In addition, it uses a Straight-Through Estimator (STE)~\citep{bengio2013estimating} to propagate gradients through the non-differentiable rounding operation, enabling end-to-end training under simulated quantization.

\subsection{Loss Function for the Reranking Model}
We frame reranking as a binary classification problem: given a query–document pair, the model predicts either a special yes token (relevant) or no token (irrelevant).

\begin{equation}
\label{loss:reranking}
        \mathcal{L}_\textrm{reranking} = -\log p(l|I,q,d),
\end{equation}

where $p(\cdot|*)$ denotes the probability assigned by the VLM. The label $l$ is ``yes'' for positive pairs and ``no'' for negatives. This loss function encourages the model to assign higher probabilities to correct labels, thereby improving the ranking performance~\citep{dai2025supervised}.

During inference, the final relevance score is computed by applying the sigmoid function to the difference between the logits of the 'yes' and 'no' tokens:
\begin{equation}
    s = \text{sigmoid}(\text{logit}(\text{yes}) - \text{logit}(\text{no})).
\end{equation}

\section{Evaluation}

\subsection{Multimodal Benchmarks}

\begin{table}[!t]
	\centering
	\renewcommand{\arraystretch}{1.2}
	\huge
	\caption{
    Results on the MMEB-V2 benchmark~\citep{meng2025vlm2vec}. CLS: classification, QA: question answering, RET: retrieval, GD: grounding, MRET: moment retrieval, VDR: ViDoRe, VR: VisRAG, OOD: out-of-distribution. $\textsuperscript{\dag}$: link to the model's homepage.
    All models except IFM-TTE have been re-evaluated on the updated VisDoc OOD\footnotemark split.
    }
\label{tab:mmeb-v2}
\resizebox{\textwidth}{!}{
	\begin{tabular}{l c ccccc ccccc ccccc c}
		\toprule
		\multirow{2}{*}{\textbf{Model}} 
		& \multirow{2}{*}{\textbf{Size}}
		& \multicolumn{5}{c}{\textbf{Image}} 
		& \multicolumn{5}{c}{\textbf{Video}} 
		& \multicolumn{5}{c}{\textbf{VisDoc}}
		& \multirow{2}{*}{\textbf{All}}\\
		\cmidrule(lr){3-7} \cmidrule(lr){8-12} \cmidrule(lr){13-17}
		& & \textbf{CLS} & \textbf{QA} & \textbf{RET} & \textbf{GD} & \textbf{Overall} 
		& \textbf{CLS} & \textbf{QA} & \textbf{RET} & \textbf{MRET} & \textbf{Overall} 
		& \textbf{VDRv1} & \textbf{VDRv2} & \textbf{VR} & \textbf{OOD} & \textbf{Overall} \\
		\midrule
		\textbf{\# of Datasets} $\rightarrow$ &
		& 10 & 10 & 12 & 4 & 36 
		& 5 & 5 & 5 & 3 & 18 
		& 10 & 4 & 6 & 4 & 24
		& 78 
		\\

        \midrule
		\multicolumn{17}{c}{\textbf{\emph{Open-Source Models}}} \\
		\midrule
		
		VLM2Vec~\citep{jiang2025vlm2vec} 
        & 2B &
        58.7 & 49.3 & 65.0 & 72.9 & 59.7 & 
        33.4 & 30.5 & 20.6 & 30.7 & 28.6 & 
        49.8 & 13.5 & 51.8 & 48.2 & 44.0 & 
        47.7 \\
		VLM2Vec-V2~\citep{meng2025vlm2vec} 
        & 2B &
        62.9 & 56.3 & 69.5 & 77.3 & 64.9 & 
        39.3 & 34.3 & 28.8 & 36.8 & 34.6 & 
        75.5 & 44.9 & 79.4 & 62.2 & 69.2 & 
        59.2 \\
		
		GME~\citep{zhang2025bridging} 
        & 2B  &
        54.4 & 29.9 & 66.9 & 55.5 & 51.9 & 
        34.9 & 42.0 & 25.6 & 31.1 & 33.6 & 
        86.1 & 54.0 & 82.5 & 67.5 & 76.8 & 
        55.3 \\

		Ops-MM-embedding-v1\href{https://huggingface.co/OpenSearch-AI/Ops-MM-embedding-v1-2B}{$\textsuperscript{\dag}$} 
        & 2B &
        68.1 & 65.1 & 69.2 & 80.9 & 69.0 & 
        53.6 & 55.6 & 41.8 & 33.7 & 47.6 & 
        76.4 & 53.2 & 77.6 & 64.2 & 70.8 & 
        64.6 \\
		
		RzenEmbed~\citep{jian2025rzenembed} 
        & 2B  &
		68.5& 66.3 & 74.5 & 90.3 & 72.3 & 
		50.4& 49.7 & 46.6 & 38.9 & 47.3 & 
		87.1& 55.1 & 87.2 & 43.4 & 74.5 & 67.2
		\\
		
		\cmidrule{1-18}
		
		VLM2Vec~\citep{jiang2025vlm2vec} 
        & 8B &
        62.7 & 56.9 & 69.4 & 82.2 & 65.5 & 
        39.1 & 30.0 & 29.0 & 38.9 & 33.7 & 
        56.9 &  9.4 & 59.1 & 54.0 & 49.1 & 
        53.1 \\
		
		GME~\citep{zhang2025bridging} 
        & 8B  &
        57.7 & 34.7 & 71.2 & 59.3 & 56.0 & 
        37.4 & 50.4 & 28.4 & 37.0 & 38.4 & 
        89.4 & 55.6 & 85.0 & 68.3 & 79.3 & 
        59.1 \\
		
		Ops-MM-embedding-v1\href{https://huggingface.co/OpenSearch-AI/Ops-MM-embedding-v1-7B}{$\textsuperscript{\dag}$} 
        & 8B &
        69.7 & 69.6 & 73.1 & 87.2 & 72.7 & 
        59.7 & 62.2 & 45.7 & 43.2 & 53.8 & 
        80.1 & 59.6 & 79.3 & 67.8 & 74.4 & 
        68.9 \\
		
		RzenEmbed~\citep{jian2025rzenembed} 
        & 8B  &
        70.6 & 71.7 & 78.5 & 92.1 & 75.9 & 
        58.8 & 63.5 & 51.0 & 45.5 & 55.7 & 
        89.7 & 60.7 & 88.7 & 69.9 & 81.3 & 
        72.9 \\

        \midrule
        \multicolumn{17}{c}{\textbf{\emph{Closed-Source Models}}} \\ 
        \midrule
        IFM-TTE\href{https://interestfm-tte.github.io/}{$\textsuperscript{\dag}$} 
                    & 8B &
					76.7 & 78.5 & 74.6 & 89.3 & 77.9 & 
					60.5 & 67.9 & 51.7 & 54.9 & 59.2 & 
					85.2 & 71.5 & 92.7 & 53.3 & 79.5 & 74.1\\

        Seed-1.6-embedding-0615\href{https://seed1-6-embedding.github.io/}{$\textsuperscript{\dag}$} & - &
					76.1 & 74.0 & 77.9 & 91.3 & 77.8 & 
					55.0 & 60.8 & 51.3 & 53.5 & 55.3 & 
					85.3 & 56.6 & 84.7 & 68.6 & 77.7 & 72.6 \\

        Seed-1.6-embedding-1215\href{https://seed1-6-embedding.github.io/}{$\textsuperscript{\dag}$} & - &
                    75.0 & 74.9 & 79.3 & 89.0 & 78.0 & 
                    85.2 & 66.7 & 59.1 & 54.8 & 67.7 & 
                    90.0 & 60.3 & 90.0 & 70.7 & 82.2 & 
                    76.9 \\
        
		\midrule
		\multicolumn{17}{c}{\textbf{\emph{Qwen3 VL Embedding Models}}} \\
		\midrule
		
		\textbf{Qwen3-VL-Embedding-2B} 
        & 2B &
		70.3 & 74.3 & 74.8 & 88.5 & 75.0 & 
		71.9 & 64.9 & 53.9 & 53.3 & 61.9 & 
		84.4 & 65.3 & 86.4 & 69.4 & 79.2 & 
		73.2    \\
		
		\textbf{Qwen3-VL-Embedding-8B} 
        & 8B &
		74.2 & 81.1 & 80.2 & 92.3 & 80.1 & 
		78.4 & 71.0 & 58.7 & 56.1 & 67.1 & 
		87.2 & 69.9 & 88.7 & 73.3 & 82.4 & 
		77.8    \\
		
		\bottomrule
	\end{tabular}
}
\end{table}
\footnotetext{
    \url{https://huggingface.co/datasets/VLM2Vec/MMLongBench-page-fixed}, \\
    \hspace*{1.8em}\url{https://huggingface.co/datasets/VLM2Vec/ViDoSeek-page-fixed}
}

To evaluate the overall performance of Qwen3-VL-Embedding in multimodal and multi-task representation learning, we report its results on the MMEB-v2 benchmark~\citep{meng2025vlm2vec}. MMEB-v2 provides a comprehensive assessment spanning three primary domains—Image, Video, and Visual Document—comprising nine task categories and 78 datasets in total. We compared our model against several prominent open-source and proprietary baselines. During evaluation, the context length is constrained to 16,384 tokens. For image-based tasks, the maximum token consumption is set at 1,800, while for video-based tasks, we cap the total tokens at 15,000 and the frame count at 64. As summarized in Table~\ref{tab:mmeb-v2}, the results demonstrate that our model achieves state-of-the-art (SOTA) average performance and exhibits exceptional proficiency across all three domains. Specifically, Qwen3-VL-Embedding-8B achieves an average score of 77.8 on MMEB-v2, representing a 6.7\% improvement over the previous best open-source model.

\subsection{Visual Document Benchmarks}

In addition to the evaluation datasets in MMEB-V2, we conducted further tests on the latest JinaVDR~\citep{günther2025jinaembeddingsv4universalembeddingsmultimodal} and Vidore-v3~\footnote{\url{https://huggingface.co/collections/vidore/vidore-benchmark-v3}} benchmarks for visual document retrieval tasks. We compared our models with current state-of-the-art ColPali-style models, with the results illustrated in Table~\ref{tab:visdoc}. As shown, our embedding model achieves performance comparable to ColPali-style models that require significantly higher computational costs. Furthermore, our reranker model substantially outperforms ColPali models of a similar parameter size.

\begin{table}[!t]
    \centering
    \renewcommand{\arraystretch}{1.2}     
    \huge
    \caption{Results on visual document retrieval benchmarks. All results are obtained from our experimental runs.}
    \label{tab:visdoc}
    \resizebox{\textwidth}{!}{%
    \begin{tabular}{lcccccccc}
        \toprule
        Model & Size & VisRAG & VisDocOOD & Vidore-v1 & Vidore-v2 & Vidore-v3 & JinaVDR & Avg \\ 
        \midrule
        llama-nemoretriever-colembed-1b-v1~\citep{xu2025llamanemoretrievercolembedtopperforming} & 1B & 82.4 & 65.6 & 90.5 & 62.1 & 55.5 & 66.4 & 70.4 \\
        llama-nemoretriever-colembed-3b-v1~\citep{xu2025llamanemoretrievercolembedtopperforming} & 3B & 85.5 & 69.7 & 91.0 & 55.5 & 57.1 & 67.8 & 71.1 \\
        colnomic-embed-multimodal-3b~\citep{nomicembedmultimodal2025} & 3B & 86.8 & 71.0 & 89.7 & 63.5 & 56.4 & 77.6 & 74.2 \\
        colqwen2.5-v0.2~\citep{faysse2025colpali} & 3B & 86.6 & 70.9 & 89.5 & 59.3 & 52.4 & 75.6 & 72.4 \\ 
        \midrule
        tomoro-colqwen3-embed-4b~\citep{huang2025beyond} & 4B & 89.0 & 75.9 & 90.6 & 66.0 & 60.2 & 76.2 & 76.5 \\
        colnomic-embed-multimodal-7b~\citep{nomicembedmultimodal2025} & 7B & 88.7 & 75.6 & 90.0 & 62.0 & 57.6 & 78.9 & 75.5 \\
        tomoro-colqwen3-embed-8b~\citep{huang2025beyond} & 8B & 90.2 & 76.8 & 90.8 & 67.7 & 61.6 & 79.2 & 77.7 \\
        \midrule
        \multicolumn{9}{c}{\textbf{\emph{Qwen3 VL Embedding Models}}} \\
        \midrule
        Qwen3-VL-Embedding-2B & 2B & 86.4 & 69.4 & 84.4 & 65.3 & 52.9 & 71.0 & 71.6 \\
        Qwen3-VL-Embedding-8B & 8B & 88.7 & 73.3 & 87.2 & 69.9 & 59.0 & 76.9 & 75.8 \\ 
        \midrule
        \multicolumn{9}{c}{\textbf{\emph{Qwen3 VL Reranking Models}}} \\
        \midrule
        Qwen3-VL-Ranker-2B & 2B & 90.2 & 72.5 & 90.5 & 65.2 & 60.8 & 80.9 & 76.7 \\
        Qwen3-VL-Ranker-8B & 8B & 91.2 & 75.7 & 91.9 & 72.8 & 66.7 & 83.6 & 80.3 \\ 
        \bottomrule
    \end{tabular}%
    }
\end{table}

\subsection{Text Benchmarks}

Table~\ref{tab:mmteb} compares our Qwen3-VL-Embedding models with standard text-only embedding models on the MMTEB~\citep{enevoldsen2025mmtebmassivemultilingualtext} benchmark. Compared to text-only Qwen3 embedding models of similar sizes, the Qwen3-VL-Embedding model series show slightly lower performance. Nevertheless, Qwen3-VL-Embedding maintains competitive performance on pure text tasks. Specifically, Qwen3-VL-Embedding-8B achieves a mean task score of 67.9 on MMTEB, performing on par with other similarly sized text-only embedding models.

\begin{table}[!t]
    \centering
    \renewcommand{\arraystretch}{1.2}
    \caption{Performance on MTEB Multilingual~\citep{enevoldsen2025mmteb}. For compared models, the scores are retrieved from MTEB online leaderboard on December 25th, 2025.}
    \label{tab:mmteb}
    \resizebox{\textwidth}{!}{
        \begin{tabular}{lc|cc|ccccccccc}
            \hline
            \textbf{Model} & \textbf{Size} & \textbf{\begin{tabular}{@{}c@{}}Mean\\(Task)\end{tabular}} & \textbf{\begin{tabular}{@{}c@{}}Mean\\(Type)\end{tabular}} & \textbf{\begin{tabular}{@{}c@{}}Bitext\\Mining\end{tabular}} & \textbf{\begin{tabular}{@{}c@{}}Class-\\ification\end{tabular}} & \textbf{\begin{tabular}{@{}c@{}}Clus-\\tering\end{tabular}} & \textbf{\begin{tabular}{@{}c@{}}Inst.\\Retrieval\end{tabular}} & \textbf{\begin{tabular}{@{}c@{}}Multilabel\\Class.\end{tabular}} & \textbf{\begin{tabular}{@{}c@{}}Pair\\Class.\end{tabular}} & \textbf{Rerank} & \textbf{Retrieval} & \textbf{STS} \\
            \hline
            \multicolumn{13}{c}{\textbf{\emph{Open-Source Models}}} \\
            \hline
            KaLM-Embedding-Gemma3-12B-2511~\citep{zhao2025kalmembeddingv2} & 12B & 72.3 & 62.5 & 83.8 & 77.9 & 55.8 & 5.5 & 33.0 & 84.7 & 67.3 & 75.7 & 79.0 \\
            llama-embed-nemotron-8b~\citep{babakhin2025llamaembednemotron8buniversaltextembedding}  & 8B & 69.5 & 61.1 & 81.7 & 73.2 & 54.4 & 10.8 & 29.9 & 84.0 & 67.8 & 68.7 & 79.4 \\
            NV-Embed-v2~\cite{lee2024nv} & 7B & 56.3 & 49.6 & 57.8 & 57.3 & 40.8 & 1.0 & 18.6 & 78.9 & 63.8 & 56.7 & 71.1 \\
            GritLM-7B~\citep{muennighoff2024generative} & 7B & 60.9 & 53.7 & 70.5 & 61.8 & 49.8 & 3.5 & 22.8 & 80.9 & 63.8 & 58.3 & 73.3 \\
            BGE-M3~\citep{chen-etal-2024-m3} & 0.6B & 59.6 & 52.2 & 79.1 & 60.4 & 40.9 & -3.1 & 20.1 & 80.8 & 62.8 & 54.6 & 74.1 \\
            multilingual-e5-large-instruct~\citep{wang2024multilingual} & 0.6B & 63.2 & 55.1 & 80.1 & 64.9 & 50.8 & -0.4 & 22.9 & 80.9 & 62.6 & 57.1 & 76.8 \\
            gte-Qwen2-1.5B-instruct~\citep{li2023towards} & 1.5B & 59.5 & 52.7 & 62.5 & 58.3 & 52.1 & 0.74 & 24.0 & 81.6 & 62.6 & 60.8 & 71.6 \\
            gte-Qwen2-7b-instruct~\citep{li2023towards} & 7B & 62.5 & 55.9 & 73.9 & 61.6 & 52.8 & 4.9 & 25.5 & 85.1 & 65.6 & 60.1 & 74.0 \\
            Qwen3-Embedding-0.6B~\citep{qwen3embedding} & 0.6B & 64.3 & 56.0 & 72.2 & 66.8 & 52.3 & 5.1 & 24.6 & 80.8 & 61.4 & 64.6 & 76.2 \\
            Qwen3-Embedding-4B~\citep{qwen3embedding} & 4B & 69.5 & 60.9 & 79.4 & 72.3 & 57.2 & 11.6 & 26.8 & 85.1 & 65.1 & 69.6 & 80.9 \\
            Qwen3-Embedding-8B~\citep{qwen3embedding} & 8B & 70.6 & 61.7 & 80.9 & 74.0 & 57.7 & 10.1 & 28.7 & 86.4 & 65.6 & 70.9 & 81.1 \\ 
            \hline
            \multicolumn{13}{c}{\textbf{\emph{Closed-Source Models}}} \\
            \hline
            text-embedding-3-large~\href{https://platform.openai.com/docs/models/text-embedding-3-large}{$\textsuperscript{\dag}$} & - & 58.9 & 51.4 & 62.2 & 60.3 & 46.9 & -2.7 & 22.0 & 79.2 & 63.9 & 59.3 & 71.7 \\
            Cohere-embed-multilingual-v3.0~\href{https://cohere.com/blog/introducing-embed-v3}{$\textsuperscript{\dag}$} & - & 61.1 & 53.2 & 70.5 & 63.0 & 46.9 & -1.9 & 22.7 & 79.9 & 64.1 & 59.2 & 74.8 \\
            Gemini Embedding~\citep{lee2025gemini} & - & 68.4 & 59.6 & 79.3 & 71.8 & 54.6 & 5.2 & 29.2 & 83.6 & 65.6 & 67.7 & 79.4 \\
            Seed-1.6-embedding-1215\href{https://seed1-6-embedding.github.io/}{$\textsuperscript{\dag}$} & - & 70.3 & 61.3 & 78.7 & 76.8 &  56.8 & -0.0 & 46.2 & 85.5 & 66.2 & 66.1 & 75.9 \\
            \hline
            \multicolumn{13}{c}{\textbf{\emph{Qwen3 VL Embedding Models}}} \\
            \hline
            Qwen3-VL-Embeddnig-2B & 2B & 63.9 & 55.8 & 69.5 & 65.9 & 52.5 & 3.9 & 26.1 & 78.5 & 64.8 & 67.1 & 74.3 \\
            Qwen3-VL-Embeddnig-8B & 8B & 67.9 & 58.9 & 77.5 & 72.0 & 55.8 & 4.5 & 28.6 & 81.1 & 65.7 & 69.4 & 75.4 \\
            \hline
        \end{tabular}
    }
\end{table}

\subsection{Evaluation for Reranking Model}

Table~\ref{tab:rerank_results} presents the evaluation results across various reranking tasks. For multimodal retrieval, we utilize the MMEB-v2 suite, covering image, video (including moment retrieval), and visual document tasks. Text retrieval is evaluated using MMTEB, while visual document retrieval is further assessed on MMEB-v2, JinaVDR, and ViDoRe v3. To ensure a fair comparison, we use Qwen3-VL-Embedding-2B to retrieve the top 100 candidates before applying the reranking models for refinement. Our results demonstrate that all three Qwen3-VL-Reranker models consistently outperform the base embedding model and baseline rerankers, with the 8B variant achieving the best performance across most tasks.

\begin{table}[t]
    \centering
    \renewcommand{\arraystretch}{1.2}
    \huge
    \caption{Evaluation results for reranking models and baselines. All scores are obtained from our experimental runs.}
    \label{tab:rerank_results}
    \resizebox{\textwidth}{!}{
    \begin{tabular}{lccccccccc}
        \toprule
        \multirow{2}{*}{\textbf{Model}} & \multirow{2}{*}{\textbf{Size}} & \multicolumn{4}{c}{\textbf{MMEB-v2(Retrieval)}} & \multirow{2}{*}{\textbf{MMTEB(Retrieval)}} & \multirow{2}{*}{\textbf{JinaVDR}} & \multirow{2}{*}{\textbf{ViDoRe(v3)}} \\
        \cmidrule(lr){3-6}
        & & \textbf{Avg} & \textbf{Image} & \textbf{Video} & \textbf{VisDoc} & & \\
        \midrule
        Qwen3-VL-Embedding-2B & 2B & 73.4 & 74.8 & 53.6 & 79.2 & 68.1 &  71.0 & 52.9 &\\
        \midrule
        jina-reranker-m0~\href{https://huggingface.co/jinaai/jina-reranker-m0}{$\textsuperscript{\dag}$}  & 2B & - & 68.2 & - & 85.2 & - & 82.2 & 57.8 \\
        \midrule
        Qwen3-VL-Reranker-2B & 2B & 75.2 & 74.0 & 53.2 & 83.2 & 70.0 &  80.9 & 60.8  \\
        Qwen3-VL-Reranker-8B & 8B & 79.2 & 78.2 & 61.0 & 85.8 & 74.9 & 83.6 & 66.7 & \\
        \bottomrule
    \end{tabular}
    }
\end{table}

\section{Analysis}

\subsection{Efficacy of Matryoshka Representation Learning and Embedding Quantization}

\begin{figure}[htbp]
\centering
\includegraphics[width= 1\linewidth]{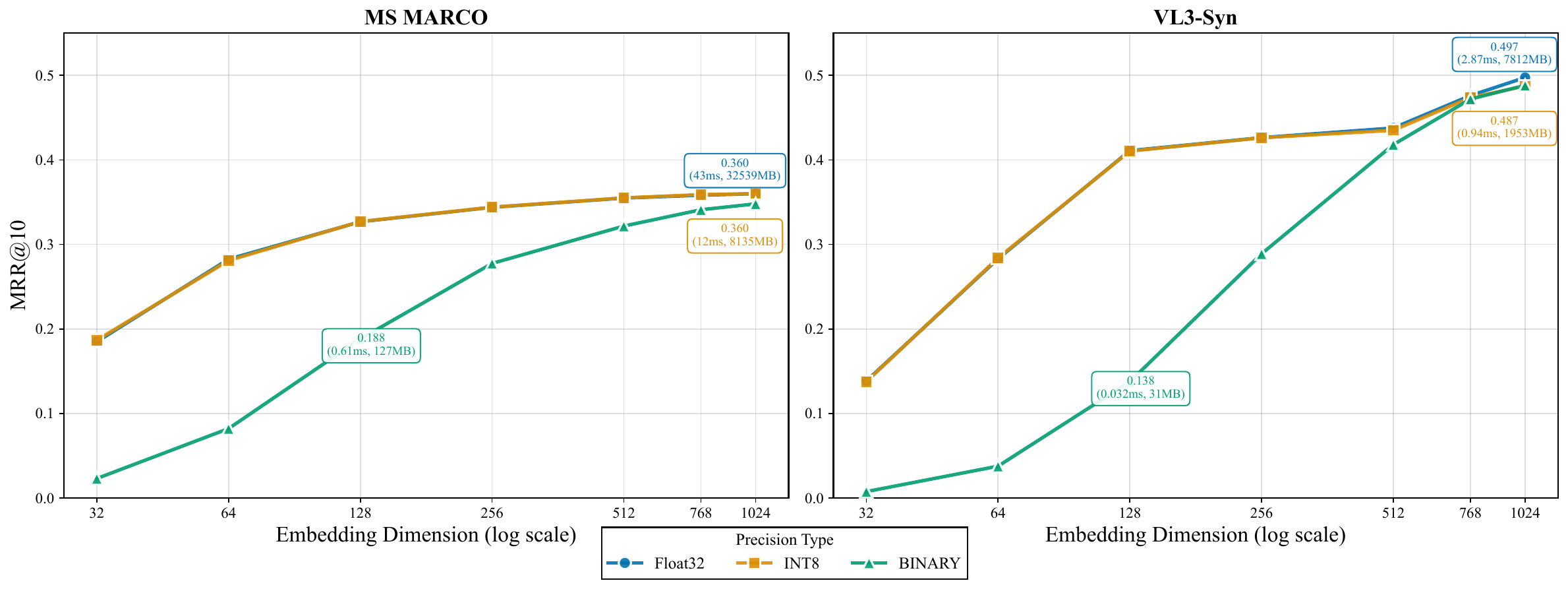}
\caption{
Performance analysis of different embedding dimensions and embedding quantization on MSMARCO Passage Dataset (text to text retrieval) and VL3-Syn Dataset (text to image retrieval).
}
\label{fig:analysis_mrl_qat}
\end{figure}

Embedding models are foundational to modern retrieval systems, spanning both unimodal tasks (e.g., text retrieval) and cross-modal scenarios (e.g., text to image retrieval). In large-scale production environments, the corpus size often reaches millions or even billions of entries. Consequently, optimizing storage requirements for the corpus while enhancing computational efficiency by reducing retrieval latency is a critical challenge. The Qwen3-VL-Embedding series addresses these needs by integrating Matryoshka Representation Learning (MRL) and Quantization-Aware Training (QAT) into its training pipeline.

To evaluate the practical impact of these strategies on retrieval performance, we conduct benchmarks across two representative tasks. The first is a text retrieval task utilizing the MSMARCO Passage Ranking dataset~\citep{msmarco}, where we sample 10,000 queries and use all passages from the training dataset as our test corpus. The second is a cross-modal text to image retrieval task based on the VL3-Syn~\citep{videollama3} dataset, featuring 10,000 captions as queries and a corpus of 2,000,000 images. We adopt the Qwen3-VL-Embedding-2B model for experimentation and utilize MRR@10 as our primary evaluation metric. Furthermore, we provide a comprehensive analysis of index storage overhead and retrieval latency across varying embedding dimensions and quantization schemes to demonstrate the tradeoffs between accuracy and efficiency.

As illustrated in Figure~\ref{fig:analysis_mrl_qat}, we observe consistent patterns in both text retrieval and text to image cross-modal retrieval. Regarding embedding dimensionality, retrieval performance degrades as dimensions decrease; however, within a reasonable range, this degradation is acceptable given the substantial savings in storage and retrieval latency. For instance, in text retrieval tasks, reducing the embedding dimension from 1024 to 512 results in only a 1.4\% decrease in retrieval performance while achieving 50\% storage reduction and doubling retrieval speed. Regarding embedding quantization, we find that int8 quantization preserves retrieval performance with negligible degradation, whereas binary quantization significantly impairs retrieval effectiveness. Moreover, this performance loss becomes increasingly pronounced as embedding dimensionality decreases.

\subsection{Impact of Spatial and Temporal Granularity}

\begin{figure}[htbp]
    \centering
    \includegraphics[width= 1\linewidth]{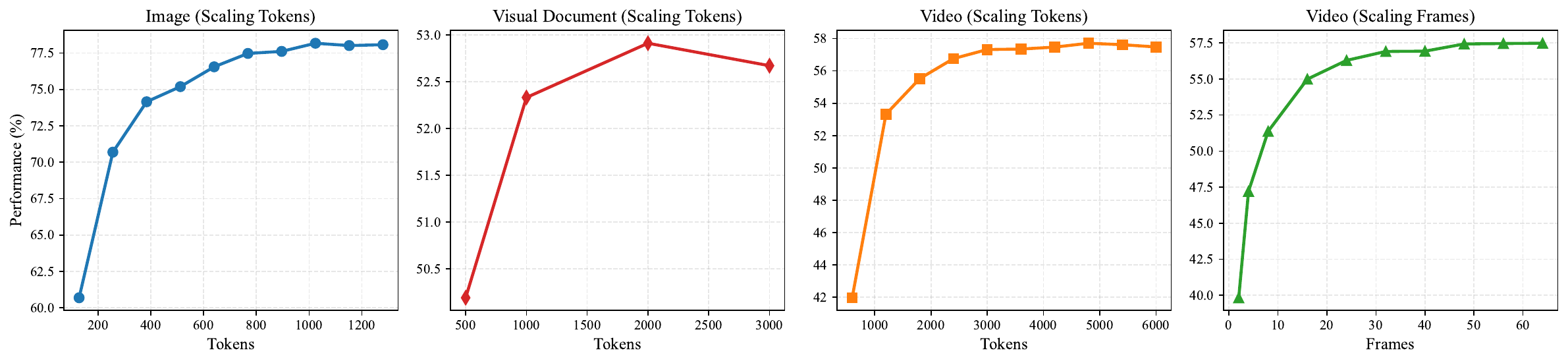}
    \caption{
       Impact of visual granularity on model performance across different domains.
    }
\label{fig:ablation_budget}
\end{figure}

In this section, we investigate how model performance scales with visual granularity across different dimensions. Specifically, for the image modality, we examine the impact of spatial resolution as measured by the number of visual tokens. For video, we decouple our analysis into two axes: (i) temporal granularity, measured by the number of frames, and (ii) spatial resolution, quantified by the aggregate token budget across all frames. 

We first analyzed the distribution of image/video resolutions and frame counts across the MMEB-v2 benchmarks, selecting several high-resolution tasks from the Image, Video, and Visual Document domains for our experiments. The results are illustrated in Figure~\ref{fig:ablation_budget}. Our findings indicate a consistent trend where performance improves with increased resource consumption across all task categories. However, we observe a pronounced diminishing return as resource allocation grows, with a slight performance regression occurring at the highest levels of consumption. A potential explanation for this decline is the inherent performance degradation that the model encounters when processing excessively long contexts.

\subsection{Performance Across Training Stages}

\begin{table}[!t]
	\centering
	\renewcommand{\arraystretch}{1.2}
	\huge
	\caption{
    Performance of Qwen3-VL-Embedding-2B across different training stages on the MMEB-V2.
    }
\label{tab:stage_performance}
\resizebox{\textwidth}{!}{
\begin{tabular}{l ccccc ccccc ccccc c}
    \toprule
    \multirow{2}{*}{
    \begin{tabular}{@{}l@{}}Model\\Stage\end{tabular} 
    } & \multicolumn{5}{c}{\textbf{Image}} & \multicolumn{5}{c}{\textbf{Video}} & \multicolumn{5}{c}{\textbf{VisDoc}} & \multirow{2}{*}{\textbf{All}} \\
    \cmidrule(lr){2-6} \cmidrule(lr){7-11} \cmidrule(lr){12-16}
    & CLS & QA & RET & GD & Overall & CLS & QA & RET & MRET & Overall & VDRv1 & VDRv2 & VR & OOD & Overall & \\
    \midrule
    s0 & 62.2 & 63.7 & 65.9 & 80.0 & 65.8 & 60.8 & 65.9 & 51.1 & 48.4 & 57.5 & 76.7 & 59.8 & 79.5 & 64.3 & 74.8 & 66.6 \\
    s1 & 71.2 & 75.8 & 72.4 & 88.3 & 74.8 & 73.0 & 67.7 & 51.3 & 41.6 & 60.3 & 83.5 & 58.8 & 84.9 & 66.4 & 77.1 & 72.1 \\
    s2 & 61.8 & 69.8 & 78.8 & 76.3 & 71.3 & 63.9 & 60.0 & 55.6 & 57.8 & 59.5 & 84.2 & 72.4 & 87.9 & 70.6 & 80.9 & 71.5 \\
    s3 & 70.3 & 74.3 & 74.8 & 88.5 & 75.0 & 71.9 & 64.9 & 53.9 & 53.3 & 61.9 & 84.4 & 65.3 & 86.4 & 69.4 & 79.2 & 73.2 \\
    \bottomrule
\end{tabular}
}
\end{table}

In our multi-stage training pipeline, a total of four embedding models were produced. Table~\ref{tab:stage_performance} details the performance of these four models at the 2B size. The results indicate that by distilling from a reranking model, the embedding model achieves a substantial performance boost in retrieval-oriented tasks. Although a slight decline is observed in other task categories during this process, the final model merging stage successfully reconciles these trade-offs, leading to a robust and superior overall performance across all benchmarks.

\section{Conclusion}
In this work, we present Qwen3-VL-Embedding and Qwen3-VL-Reranker, a state-of-the-art model series for multimodal retrieval. By integrating a multi-stage training pipeline with high-quality multimodal data while maximally leveraging the multimodal knowledge and general understanding capabilities of Qwen3-VL Foundation models, Qwen3-VL-Embedding and Qwen3-VL-Reranker model series achieve unprecedented performance across a broad spectrum of multimodal retrieval benchmarks while maintaining strong pure-text capabilities. Furthermore, through matryoshka representation learning and quantization-aware training, the Qwen3-VL-Embedding series offers excellent practical deployment characteristics, significantly reducing computational costs for downstream tasks while preserving superior performance. Looking forward, promising directions include extending support to additional modalities, developing more efficient training paradigms, enhancing compositional reasoning capabilities, and establishing more comprehensive evaluation protocols. We believe these models represent a significant advancement in multimodal retrieval technology and hope they will facilitate further innovation in this rapidly evolving field.

\bibliography{colm2024_conference}

@inproceedings{jiang2025vlm2vec,
  title={VLM2Vec: Training Vision-Language Models for Massive Multimodal Embedding Tasks},
  author={Jiang, Ziyan and Meng, Rui and Yang, Xinyi and Yavuz, Semih and Zhou, Yingbo and Chen, Wenhu},
  booktitle={ICLR},
  year={2025}
}

@article{10.1109/TASLP.2024.3402087,
author = {Huang, Xiang and Peng, Hao and Zou, Dongcheng and Liu, Zhiwei and Li, Jianxin and Liu, Kay and Wu, Jia and Su, Jianlin and Yu, Philip S.},
title = {CoSENT: Consistent Sentence Embedding via Similarity Ranking},
year = {2024},
issue_date = {2024},
publisher = {IEEE Press},
volume = {32},
issn = {2329-9290},
url = {https://doi.org/10.1109/TASLP.2024.3402087},
doi = {10.1109/TASLP.2024.3402087},
journal = {IEEE/ACM Trans. Audio, Speech and Lang. Proc.},
month = may,
pages = {2800–2813},
numpages = {14}
}

@misc{xu2025llamanemoretrievercolembedtopperforming,
      title={Llama Nemoretriever Colembed: Top-Performing Text-Image Retrieval Model}, 
      author={Mengyao Xu and Gabriel Moreira and Ronay Ak and Radek Osmulski and Yauhen Babakhin and Zhiding Yu and Benedikt Schifferer and Even Oldridge},
      year={2025},
      eprint={2507.05513},
      archivePrefix={arXiv},
      primaryClass={cs.CV},
      url={https://arxiv.org/abs/2507.05513}, 
}

@misc{nomicembedmultimodal2025,
  title={Nomic Embed Multimodal: Interleaved Text, Image, and Screenshots for Visual Document Retrieval},
  author={Nomic Team},
  year={2025},
  publisher={Nomic AI},
  url={https://nomic.ai/blog/posts/nomic-embed-multimodal},
}

@misc{huang2025beyond,
  author = {Huang, Xin and Tan, Kye Min},
  title = {Beyond Text: Unlocking True Multimodal, End-to-end RAG with Tomoro ColQwen3},
  year = {2025},
  url = {https://tomoro.ai/insights/beyond-text-unlocking-true-multimodal-end-to-end-rag-with-tomoro-colqwen3},
  publisher = {Tomoro.ai}
}

@misc{zhao2025kalmembeddingv2,
      title={KaLM-Embedding-V2: Superior Training Techniques and Data Inspire A Versatile Embedding Model}, 
      author={Xinping Zhao and Xinshuo Hu and Zifei Shan and Shouzheng Huang and Yao Zhou and Xin Zhang and Zetian Sun and Zhenyu Liu and Dongfang Li and Xinyuan Wei and Youcheng Pan and Yang Xiang and Meishan Zhang and Haofen Wang and Jun Yu and Baotian Hu and Min Zhang},
      year={2025},
      eprint={2506.20923},
      archivePrefix={arXiv},
      primaryClass={cs.CL},
      url={https://arxiv.org/abs/2506.20923}, 
}

@misc{babakhin2025llamaembednemotron8buniversaltextembedding,
      title={Llama-Embed-Nemotron-8B: A Universal Text Embedding Model for Multilingual and Cross-Lingual Tasks}, 
      author={Yauhen Babakhin and Radek Osmulski and Ronay Ak and Gabriel Moreira and Mengyao Xu and Benedikt Schifferer and Bo Liu and Even Oldridge},
      year={2025},
      eprint={2511.07025},
      archivePrefix={arXiv},
      primaryClass={cs.CL},
      url={https://arxiv.org/abs/2511.07025}, 
}

@article{lee2024nv,
  title={NV-Embed: Improved Techniques for Training LLMs as Generalist Embedding Models},
  author={Lee, Chankyu and Roy, Rajarshi and Xu, Mengyao and Raiman, Jonathan and Shoeybi, Mohammad and Catanzaro, Bryan and Ping, Wei},
  journal={arXiv preprint arXiv:2405.17428},
  year={2024}
}

@misc{muennighoff2024generative,
      title={Generative Representational Instruction Tuning}, 
      author={Niklas Muennighoff and Hongjin Su and Liang Wang and Nan Yang and Furu Wei and Tao Yu and Amanpreet Singh and Douwe Kiela},
      year={2024},
      eprint={2402.09906},
      archivePrefix={arXiv},
      primaryClass={cs.CL}
}

@article{wang2024multilingual,
  title={Multilingual E5 Text Embeddings: A Technical Report},
  author={Wang, Liang and Yang, Nan and Huang, Xiaolong and Yang, Linjun and Majumder, Rangan and Wei, Furu},
  journal={arXiv preprint arXiv:2402.05672},
  year={2024}
}

@article{li2023towards,
  title={Towards general text embeddings with multi-stage contrastive learning},
  author={Li, Zehan and Zhang, Xin and Zhang, Yanzhao and Long, Dingkun and Xie, Pengjun and Zhang, Meishan},
  journal={arXiv preprint arXiv:2308.03281},
  year={2023}
}

@article{qwen3embedding,
  title={Qwen3 Embedding: Advancing Text Embedding and Reranking Through Foundation Models},
  author={Zhang, Yanzhao and Li, Mingxin and Long, Dingkun and Zhang, Xin and Lin, Huan and Yang, Baosong and Xie, Pengjun and Yang, An and Liu, Dayiheng and Lin, Junyang and Huang, Fei and Zhou, Jingren},
  journal={arXiv preprint arXiv:2506.05176},
  year={2025}
}

@misc{günther2025jinaembeddingsv4universalembeddingsmultimodal,
      title={jina-embeddings-v4: Universal Embeddings for Multimodal Multilingual Retrieval}, 
      author={Michael Günther and Saba Sturua and Mohammad Kalim Akram and Isabelle Mohr and Andrei Ungureanu and Sedigheh Eslami and Scott Martens and Bo Wang and Nan Wang and Han Xiao},
      year={2025},
      eprint={2506.18902},
      archivePrefix={arXiv},
      primaryClass={cs.AI},
      url={https://arxiv.org/abs/2506.18902}, 
}

@misc{enevoldsen2025mmtebmassivemultilingualtext,
      title={MMTEB: Massive Multilingual Text Embedding Benchmark}, 
      author={Kenneth Enevoldsen and Isaac Chung and Imene Kerboua and Márton Kardos and Ashwin Mathur and David Stap and Jay Gala and Wissam Siblini and Dominik Krzemiński and Genta Indra Winata and Saba Sturua and Saiteja Utpala and Mathieu Ciancone and Marion Schaeffer and Gabriel Sequeira and Diganta Misra and Shreeya Dhakal and Jonathan Rystrøm and Roman Solomatin and Ömer Çağatan and Akash Kundu and Martin Bernstorff and Shitao Xiao and Akshita Sukhlecha and Bhavish Pahwa and Rafał Poświata and Kranthi Kiran GV and Shawon Ashraf and Daniel Auras and Björn Plüster and Jan Philipp Harries and Loïc Magne and Isabelle Mohr and Mariya Hendriksen and Dawei Zhu and Hippolyte Gisserot-Boukhlef and Tom Aarsen and Jan Kostkan and Konrad Wojtasik and Taemin Lee and Marek Šuppa and Crystina Zhang and Roberta Rocca and Mohammed Hamdy and Andrianos Michail and John Yang and Manuel Faysse and Aleksei Vatolin and Nandan Thakur and Manan Dey and Dipam Vasani and Pranjal Chitale and Simone Tedeschi and Nguyen Tai and Artem Snegirev and Michael Günther and Mengzhou Xia and Weijia Shi and Xing Han Lù and Jordan Clive and Gayatri Krishnakumar and Anna Maksimova and Silvan Wehrli and Maria Tikhonova and Henil Panchal and Aleksandr Abramov and Malte Ostendorff and Zheng Liu and Simon Clematide and Lester James Miranda and Alena Fenogenova and Guangyu Song and Ruqiya Bin Safi and Wen-Ding Li and Alessia Borghini and Federico Cassano and Hongjin Su and Jimmy Lin and Howard Yen and Lasse Hansen and Sara Hooker and Chenghao Xiao and Vaibhav Adlakha and Orion Weller and Siva Reddy and Niklas Muennighoff},
      year={2025},
      eprint={2502.13595},
      archivePrefix={arXiv},
      primaryClass={cs.CL},
      url={https://arxiv.org/abs/2502.13595}, 
}

@article{meng2025vlm2vec,
  title={Vlm2vec-v2: Advancing multimodal embedding for videos, images, and visual documents},
  author={Meng, Rui and Jiang, Ziyan and Liu, Ye and Su, Mingyi and Yang, Xinyi and Fu, Yuepeng and Qin, Can and Chen, Zeyuan and Xu, Ran and Xiong, Caiming and others},
  journal={arXiv preprint arXiv:2507.04590},
  year={2025}
}

@inproceedings{zhang2025bridging,
  title={Bridging Modalities: Improving Universal Multimodal Retrieval by Multimodal Large Language Models},
  author={Zhang, Xin and Zhang, Yanzhao and Xie, Wen and Li, Mingxin and Dai, Ziqi and Long, Dingkun and Xie, Pengjun and Zhang, Meishan and Li, Wenjie and Zhang, Min},
  booktitle={Proceedings of the Computer Vision and Pattern Recognition Conference},
  pages={9274--9285},
  year={2025}
}

@inproceedings{faysse2025colpali,
  title={ColPali: Efficient Document Retrieval with Vision Language Models},
  author={Faysse, Manuel and Sibille, Hugues and Wu, Tony and Omrani, Bilel and Viaud, Gautier and Hudelot, C{\'e}line and Colombo, Pierre},
  booktitle={ICLR},
  year={2025}
}

@article{jian2025rzenembed,
  title={RzenEmbed: Towards Comprehensive Multimodal Retrieval},
  author={Jian, Weijian and Zhang, Yajun and Liang, Dawei and Xie, Chunyu and He, Yixiao and Leng, Dawei and Yin, Yuhui},
  journal={arXiv preprint arXiv:2510.27350},
  year={2025}
}

@inproceedings{enevoldsen2025mmteb,
  title={MMTEB: Massive Multilingual Text Embedding Benchmark},
  author={Enevoldsen, Kenneth and Chung, Isaac and Kerboua, Imene and Kardos, M{\'a}rton and Mathur, Ashwin and Stap, David and Gala, Jay and Siblini, Wissam and Krzeminski, Dominik and Winata, Genta Indra and others},
  booktitle={International Conference on Learning Representations},
  year={2025},
  organization={International Conference on Learning Representations}
}

@inproceedings{radford2021learning,
  title={Learning transferable visual models from natural language supervision},
  author={Radford, Alec and Kim, Jong Wook and Hallacy, Chris and Ramesh, Aditya and Goh, Gabriel and Agarwal, Sandhini and Sastry, Girish and Askell, Amanda and Mishkin, Pamela and Clark, Jack and others},
  booktitle={International conference on machine learning},
  pages={8748--8763},
  year={2021},
  organization={PmLR}
}

@article{jiang2024e5,
  title={E5-v: Universal embeddings with multimodal large language models},
  author={Jiang, Ting and Song, Minghui and Zhang, Zihan and Huang, Haizhen and Deng, Weiwei and Sun, Feng and Zhang, Qi and Wang, Deqing and Zhuang, Fuzhen},
  journal={arXiv preprint arXiv:2407.12580},
  year={2024}
}

@inproceedings{zhou2025megapairs,
  title={Megapairs: Massive data synthesis for universal multimodal retrieval},
  author={Zhou, Junjie and Xiong, Yongping and Liu, Zheng and Liu, Ze and Xiao, Shitao and Wang, Yueze and Zhao, Bo and Zhang, Chen Jason and Lian, Defu},
  booktitle={Proceedings of the 63rd Annual Meeting of the Association for Computational Linguistics (Volume 1: Long Papers)},
  pages={19076--19095},
  year={2025}
}

@inproceedings{xu2016msr,
  title={Msr-vtt: A large video description dataset for bridging video and language},
  author={Xu, Jun and Mei, Tao and Yao, Ting and Rui, Yong},
  booktitle={Proceedings of the IEEE conference on computer vision and pattern recognition},
  pages={5288--5296},
  year={2016}
}

@inproceedings{xiao2021next,
  title={Next-qa: Next phase of question-answering to explaining temporal actions},
  author={Xiao, Junbin and Shang, Xindi and Yao, Angela and Chua, Tat-Seng},
  booktitle={Proceedings of the IEEE/CVF conference on computer vision and pattern recognition},
  pages={9777--9786},
  year={2021}
}

@article{soomro2012ucf101,
  title={Ucf101: A dataset of 101 human actions classes from videos in the wild},
  author={Soomro, Khurram and Zamir, Amir Roshan and Shah, Mubarak},
  journal={arXiv preprint arXiv:1212.0402},
  year={2012}
}

@article{Qwen3-VL,
      title={Qwen3-VL Technical Report}, 
      author={Shuai Bai and Yuxuan Cai and Ruizhe Chen and Keqin Chen and Xionghui Chen and Zesen Cheng and Lianghao Deng and Wei Ding and Chang Gao and Chunjiang Ge and Wenbin Ge and Zhifang Guo and Qidong Huang and Jie Huang and Fei Huang and Binyuan Hui and Shutong Jiang and Zhaohai Li and Mingsheng Li and Mei Li and Kaixin Li and Zicheng Lin and Junyang Lin and Xuejing Liu and Jiawei Liu and Chenglong Liu and Yang Liu and Dayiheng Liu and Shixuan Liu and Dunjie Lu and Ruilin Luo and Chenxu Lv and Rui Men and Lingchen Meng and Xuancheng Ren and Xingzhang Ren and Sibo Song and Yuchong Sun and Jun Tang and Jianhong Tu and Jianqiang Wan and Peng Wang and Pengfei Wang and Qiuyue Wang and Yuxuan Wang and Tianbao Xie and Yiheng Xu and Haiyang Xu and Jin Xu and Zhibo Yang and Mingkun Yang and Jianxin Yang and An Yang and Bowen Yu and Fei Zhang and Hang Zhang and Xi Zhang and Bo Zheng and Humen Zhong and Jingren Zhou and Fan Zhou and Jing Zhou and Yuanzhi Zhu and Ke Zhu},
	  journal={arXiv preprint arXiv:2511.21631},
      year={2025}
}

@inproceedings{robinson2021contrastive,
  title={CONTRASTIVE LEARNING WITH HARD NEGATIVE SAMPLES},
  author={Robinson, Joshua and Chuang, Ching-Yao and Sra, Suvrit and Jegelka, Stefanie},
  booktitle={International Conference on Learning Representations (ICLR)},
  year={2021}
}

@article{kusupati2022matryoshka,
  title={Matryoshka representation learning},
  author={Kusupati, Aditya and Bhatt, Gantavya and Rege, Aniket and Wallingford, Matthew and Sinha, Aditya and Ramanujan, Vivek and Howard-Snyder, William and Chen, Kaifeng and Kakade, Sham and Jain, Prateek and others},
  journal={Advances in Neural Information Processing Systems},
  volume={35},
  pages={30233--30249},
  year={2022}
}

@inproceedings{esser2020learned,
  title={LEARNED STEP SIZE QUANTIZATION},
  author={Esser, Steven K and McKinstry, Jeffrey L and Bablani, Deepika and Appuswamy, Rathinakumar and Modha, Dharmendra S},
  booktitle={International Conference on Learning Representations},
  year={2020}
}

@article{bengio2013estimating,
  title={Estimating or propagating gradients through stochastic neurons for conditional computation},
  author={Bengio, Yoshua and L{\'e}onard, Nicholas and Courville, Aaron},
  journal={arXiv preprint arXiv:1308.3432},
  year={2013}
}

@article{dai2025supervised,
  title={Supervised Fine-Tuning or Contrastive Learning? Towards Better Multimodal LLM Reranking},
  author={Dai, Ziqi and Zhang, Xin and Li, Mingxin and Zhang, Yanzhao and Long, Dingkun and Xie, Pengjun and Zhang, Meishan and Li, Wenjie and Zhang, Min},
  journal={arXiv preprint arXiv:2510.14824},
  year={2025}
}

@article{oord2018representation,
  title={Representation learning with contrastive predictive coding},
  author={Oord, Aaron van den and Li, Yazhe and Vinyals, Oriol},
  journal={arXiv preprint arXiv:1807.03748},
  year={2018}
}

@article{wang2022text,
  title={Text embeddings by weakly-supervised contrastive pre-training},
  author={Wang, Liang and Yang, Nan and Huang, Xiaolong and Jiao, Binxing and Yang, Linjun and Jiang, Daxin and Majumder, Rangan and Wei, Furu},
  journal={arXiv preprint arXiv:2212.03533},
  year={2022}
}

@inproceedings{chen-etal-2024-m3,
    title = "{M}3-Embedding: Multi-Linguality, Multi-Functionality, Multi-Granularity Text Embeddings Through Self-Knowledge Distillation",
    author = "Chen, Jianlyu  and
      Xiao, Shitao  and
      Zhang, Peitian  and
      Luo, Kun  and
      Lian, Defu  and
      Liu, Zheng",
    editor = "Ku, Lun-Wei  and
      Martins, Andre  and
      Srikumar, Vivek",
    booktitle = "Findings of the Association for Computational Linguistics: ACL 2024",
    month = aug,
    year = "2024",
    address = "Bangkok, Thailand",
    publisher = "Association for Computational Linguistics",
    url = "https://aclanthology.org/2024.findings-acl.137/",
    doi = "10.18653/v1/2024.findings-acl.137",
    pages = "2318--2335"
}

@article{li2024improving,
  title={Improving General Text Embedding Model: Tackling Task Conflict and Data Imbalance through Model Merging},
  author={Li, Mingxin and Nie, Zhijie and Zhang, Yanzhao and Long, Dingkun and Zhang, Richong and Xie, Pengjun},
  journal={arXiv preprint arXiv:2410.15035},
  year={2024}
}

@inproceedings{DBLP:conf/iclr/HuSWALWWC22,
  author       = {Edward J. Hu and
                  Yelong Shen and
                  Phillip Wallis and
                  Zeyuan Allen{-}Zhu and
                  Yuanzhi Li and
                  Shean Wang and
                  Lu Wang and
                  Weizhu Chen},
  title        = {LoRA: Low-Rank Adaptation of Large Language Models},
  booktitle    = {The Tenth International Conference on Learning Representations, {ICLR}
                  2022, Virtual Event, April 25-29, 2022},
  publisher    = {OpenReview.net},
  year         = {2022},
  url          = {https://openreview.net/forum?id=nZeVKeeFYf9},
  bibsource    = {dblp computer science bibliography, https://dblp.org}
}

@inproceedings{lin2014microsoft,
  title={Microsoft coco: Common objects in context},
  author={Lin, Tsung-Yi and Maire, Michael and Belongie, Serge and Hays, James and Perona, Pietro and Ramanan, Deva and Doll{\'a}r, Piotr and Zitnick, C Lawrence},
  booktitle={European conference on computer vision},
  pages={740--755},
  year={2014},
  organization={Springer}
}

@article{fu2025moon,
  title={MOON Embedding: Multimodal Representation Learning for E-commerce Search Advertising},
  author={Fu, Chenghan and Zhang, Daoze and Lin, Yukang and Nie, Zhanheng and Zhang, Xiang and Liu, Jianyu and Liu, Yueran and Guan, Wanxian and Wang, Pengjie and Xu, Jian and others},
  journal={arXiv preprint arXiv:2511.11305},
  year={2025}
}

@article{manzoor2023multimodality,
  title={Multimodality representation learning: A survey on evolution, pretraining and its applications},
  author={Manzoor, Muhammad Arslan and Albarri, Sarah and Xian, Ziting and Meng, Zaiqiao and Nakov, Preslav and Liang, Shangsong},
  journal={ACM Transactions on Multimedia Computing, Communications and Applications},
  volume={20},
  number={3},
  pages={1--34},
  year={2023},
  publisher={ACM New York, NY}
}

@article{mei2025survey,
  title={A survey of multimodal retrieval-augmented generation},
  author={Mei, Lang and Mo, Siyu and Yang, Zhihan and Chen, Chong},
  journal={arXiv preprint arXiv:2504.08748},
  year={2025}
}

@article{wang2024qwen2,
  title={Qwen2-vl: Enhancing vision-language model's perception of the world at any resolution},
  author={Wang, Peng and Bai, Shuai and Tan, Sinan and Wang, Shijie and Fan, Zhihao and Bai, Jinze and Chen, Keqin and Liu, Xuejing and Wang, Jialin and Ge, Wenbin and others},
  journal={arXiv preprint arXiv:2409.12191},
  year={2024}
}

@article{hurst2024gpt,
  title={Gpt-4o system card},
  author={Hurst, Aaron and Lerer, Adam and Goucher, Adam P and Perelman, Adam and Ramesh, Aditya and Clark, Aidan and Ostrow, AJ and Welihinda, Akila and Hayes, Alan and Radford, Alec and others},
  journal={arXiv preprint arXiv:2410.21276},
  year={2024}
}

@article{zhang2015character,
  title={Character-level convolutional networks for text classification},
  author={Zhang, Xiang and Zhao, Junbo and LeCun, Yann},
  journal={Advances in neural information processing systems},
  volume={28},
  year={2015}
}

@article{rajpurkar2016squad,
  title={Squad: 100,000+ questions for machine comprehension of text},
  author={Rajpurkar, Pranav and Zhang, Jian and Lopyrev, Konstantin and Liang, Percy},
  journal={arXiv preprint arXiv:1606.05250},
  year={2016}
}

@article{krizhevsky2009learning,
  title={Learning multiple layers of features from tiny images},
  author={Krizhevsky, Alex and Hinton, Geoffrey and others},
  year={2009},
  publisher={Toronto, ON, Canada}
}

@inproceedings{goyal2017making,
  title={Making the v in vqa matter: Elevating the role of image understanding in visual question answering},
  author={Goyal, Yash and Khot, Tejas and Summers-Stay, Douglas and Batra, Dhruv and Parikh, Devi},
  booktitle={Proceedings of the IEEE conference on computer vision and pattern recognition},
  pages={6904--6913},
  year={2017}
}

@article{msmarco,
  title={Ms marco: A human generated machine reading comprehension dataset},
  author={Bajaj, Payal and Campos, Daniel and Craswell, Nick and Deng, Li and Gao, Jianfeng and Liu, Xiaodong and Majumder, Rangan and McNamara, Andrew and Mitra, Bhaskar and Nguyen, Tri and others},
  journal={arXiv preprint arXiv:1611.09268},
  year={2016}
}

@article{videollama3,
  title={Videollama 3: Frontier multimodal foundation models for image and video understanding},
  author={Zhang, Boqiang and Li, Kehan and Cheng, Zesen and Hu, Zhiqiang and Yuan, Yuqian and Chen, Guanzheng and Leng, Sicong and Jiang, Yuming and Zhang, Hang and Li, Xin and others},
  journal={arXiv preprint arXiv:2501.13106},
  year={2025}
}

@article{lee2025gemini,
  title={Gemini embedding: Generalizable embeddings from gemini},
  author={Lee, Jinhyuk and Chen, Feiyang and Dua, Sahil and Cer, Daniel and Shanbhogue, Madhuri and Naim, Iftekhar and {\'A}brego, Gustavo Hern{\'a}ndez and Li, Zhe and Chen, Kaifeng and Vera, Henrique Schechter and others},
  journal={arXiv preprint arXiv:2503.07891},
  year={2025}
}
\bibliographystyle{colm2024_conference}

\appendix
\clearpage

\section{Dataset Examples}
\label{sec:appendix_dataset_examples}

\newcommand{\dsGraphic}[1]{
    \includegraphics[width=1.0\linewidth, height=2.5cm, keepaspectratio]{#1}
}

\newcommand{\dsBox}[1]{
    \begin{minipage}[t]{\linewidth}
    \vspace{-6pt}
    #1
    \end{minipage}
}

\newcolumntype{Z}{>{\raggedright\arraybackslash}X}


\begin{table*}[ht!]
\centering
\small
\setlength{\tabcolsep}{8pt}
\renewcommand{\arraystretch}{1.3}
\caption{Dataset format examples: Docmatix~\href{https://huggingface.co/blog/docmatix}{$\textsuperscript{\dag}$} and MS-COCO~\citep{lin2014microsoft}.}
\label{tab:appendix-dataset-examples}

\begin{tabularx}{\textwidth}{l Z}
\toprule
\rowcolor[HTML]{F5F5F5} \textbf{Dataset} & Docmatix \\
\textbf{Instruction} & Find a screenshot that relevant to the user's question. \\
\bottomrule
\end{tabularx}


\begin{tabularx}{\textwidth}{l Z p{0cm}}
\rowcolor[HTML]{F5F5F5} \multicolumn{3}{l}{\textbf{Queries ($Q_i$)}} \\
\midrule
q\_01 & \dsBox{What type of research project was announced by the Danish Cancer Society on 01/02/21?} &  \\
\bottomrule
\end{tabularx}


\begin{tabularx}{\textwidth}{l p{1.2cm} Z}
\rowcolor[HTML]{F5F5F5} \multicolumn{3}{l}{\textbf{Corpus ($C_i$)}} \\
\midrule
d\_01 & \dsBox{\dsGraphic{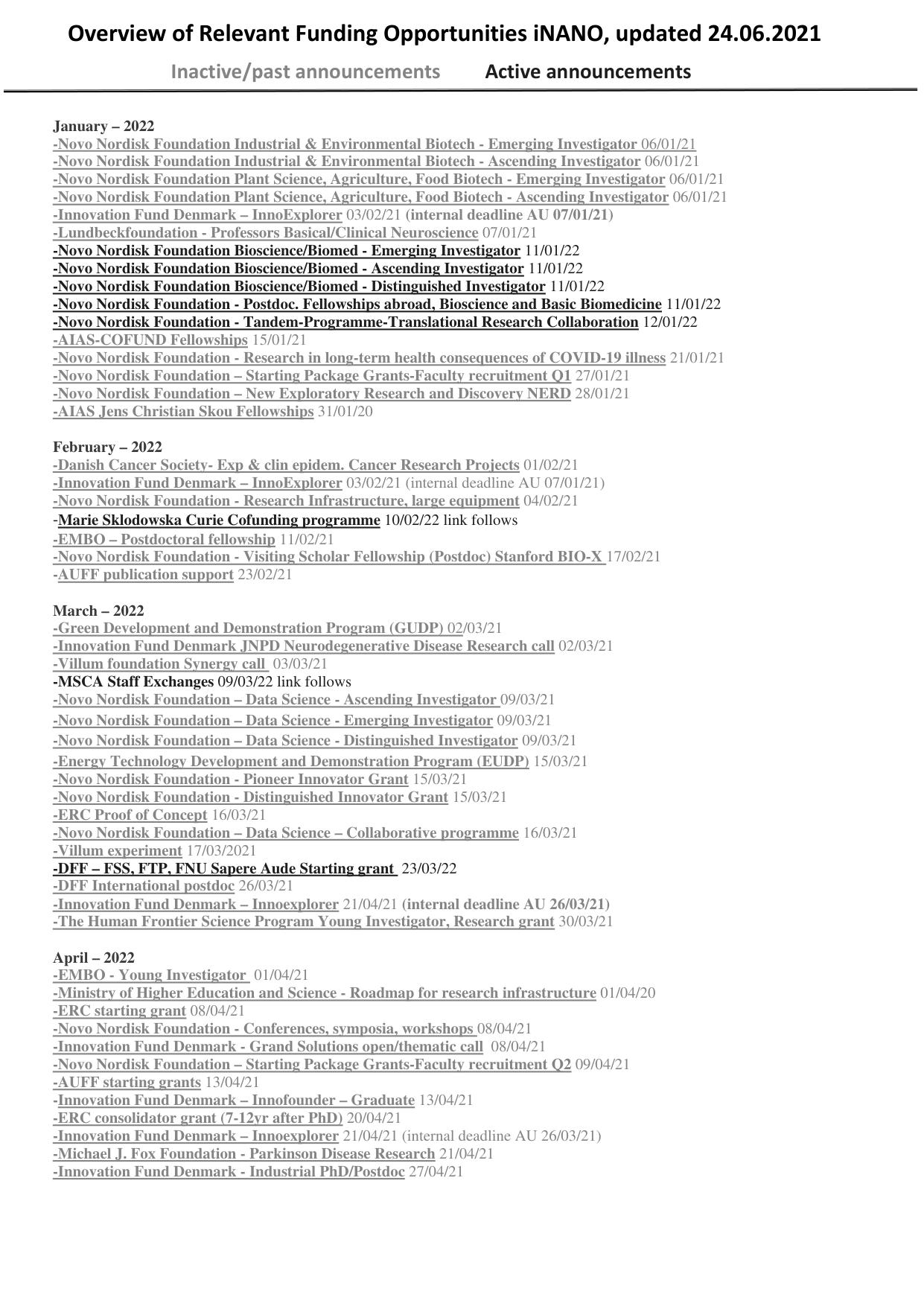}} & \\
d\_02 & \dsBox{\dsGraphic{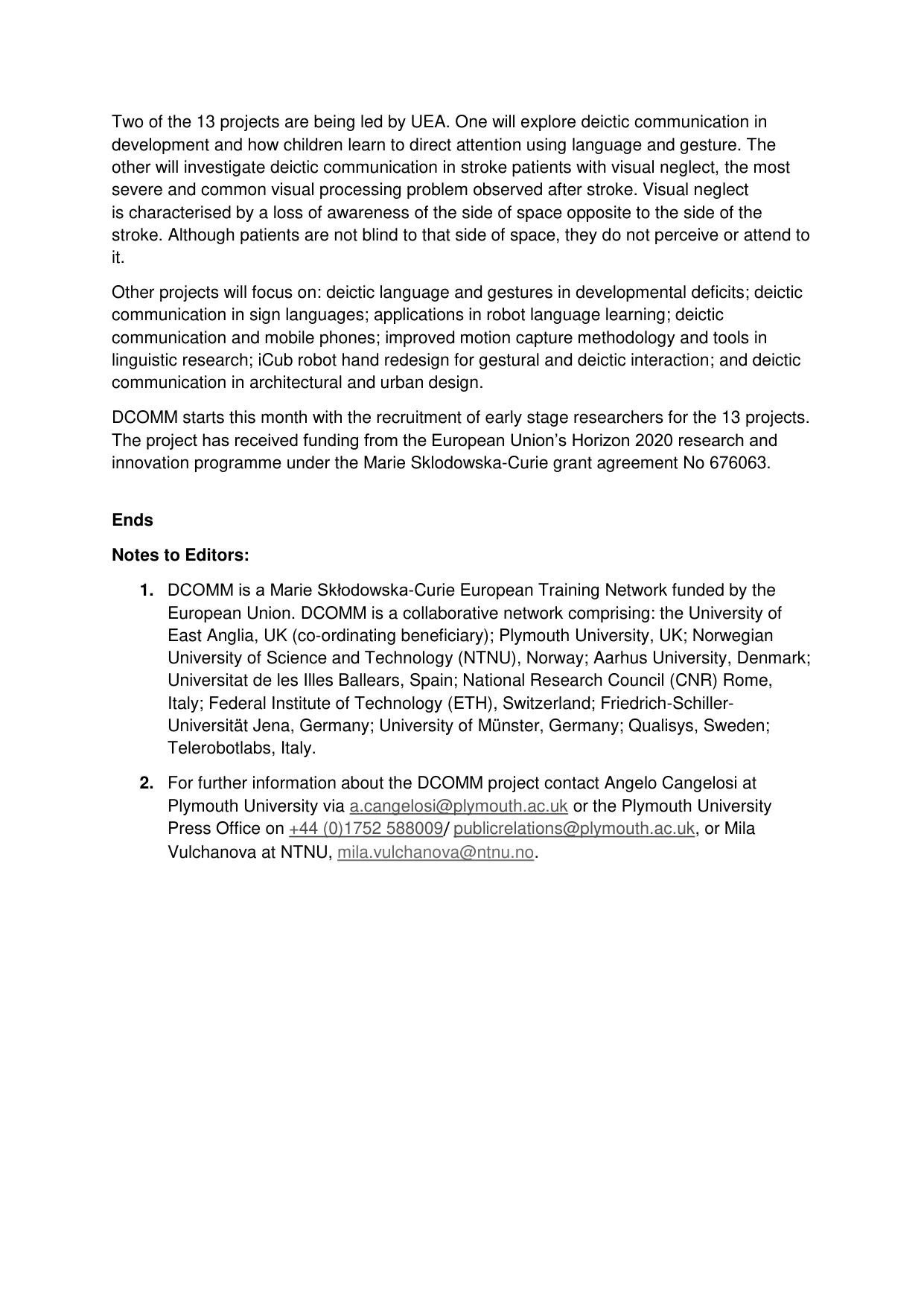}} & \\
d\_03 & \dsBox{\dsGraphic{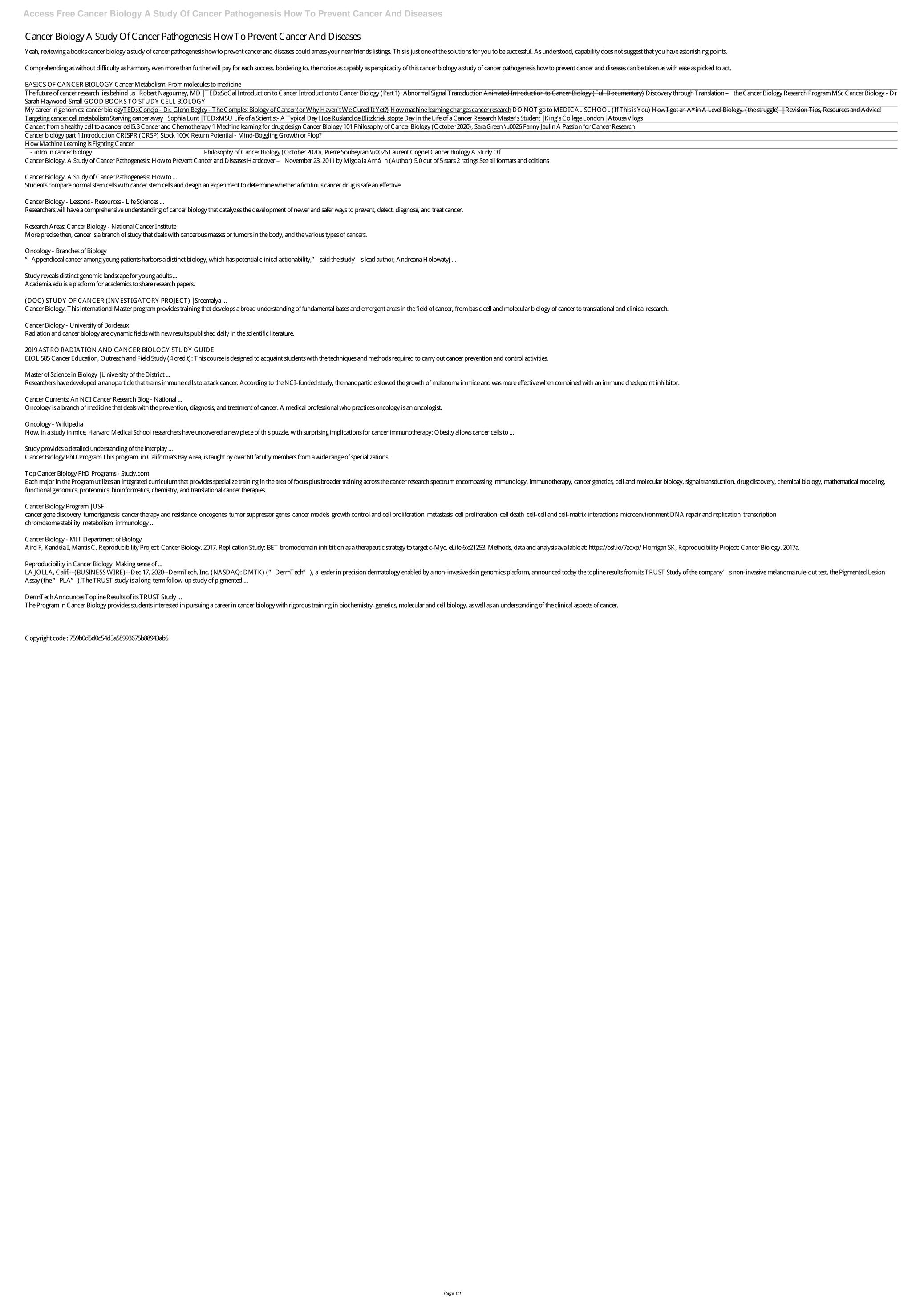}} & \\
\bottomrule
\end{tabularx}


\begin{tabularx}{\textwidth}{l Z}
\textbf{Relevance ($R_i$)} & \texttt{
\{q\_01: {pos: [d\_01]}, neg: [d\_02, d\_03]\};
} \\
\bottomrule
\end{tabularx}

\vspace{1.5em} 

\begin{tabularx}{\textwidth}{l Z}
\toprule
\rowcolor[HTML]{F5F5F5} \textbf{Dataset} & MS-COCO \\
\textbf{Instruction} & Find an image caption describing the following everyday image. \\
\bottomrule
\end{tabularx}


\begin{tabularx}{\textwidth}{l p{8cm} Z }
\rowcolor[HTML]{F5F5F5} \multicolumn{3}{l}{\textbf{Queries ($Q_i$)}} \\
\midrule
q\_01 & \dsBox{\dsGraphic{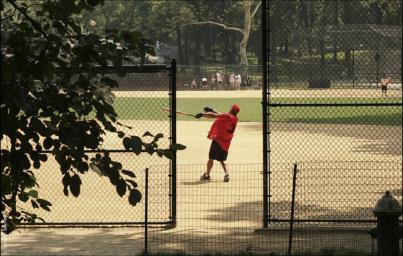}} &  \\
\bottomrule
\end{tabularx}


\begin{tabularx}{\textwidth}{l Z p{0cm}}
\rowcolor[HTML]{F5F5F5} \multicolumn{3}{l}{\textbf{Corpus ($C_i$)}} \\
\midrule
d\_01 & \dsBox{A man swinging a baseball bat on a baseball field.} &  \\
d\_02 & \dsBox{The man is walking on the field to play a game of baseball.} &  \\
d\_03 & \dsBox{A boy playing baseball waiting for a pitch.} &  \\
\bottomrule
\end{tabularx}


\begin{tabularx}{\textwidth}{l Z}
\textbf{Relevance ($R_i$)} & \texttt{
\{q\_01: {pos: [d\_01]}, neg: [d\_02, d\_03]\};
} \\
\bottomrule
\end{tabularx}

\end{table*}

\section{Examples of Data Synthesis Prompts}
\label{sec:data_synthesis_prompts}

\begin{promptblock}{Image Question Answering Prompt}
\begin{verbatim}
You are given an image. Your job is to create ONE high-quality multimodal training example for an Image Question Answering (IQA) dataset.
The final answer MUST be a single JSON object and nothing else.

STEP 1 - Visual Description (less or equal than 500 {language} words)
- General scene summary and object-level details (attributes, positions, relations).
- Contextual features (environment, lighting, actions).
- Brainstorm the types of reasoning enabled (e.g., spatial, comparative, predictive).

STEP 2 - Task Selection
Choose ONE task type from the list below that best fits the image content:
- Factoid Identification: Questions about specific entities, brands, or basic facts (e.g., "What brand is the watch?").
- Visual Reasoning: Questions requiring logical inference or analysis (e.g., "How many rats were fed the control diet?").
- OCR-based Data Extraction: Questions targeting text, tables, or document info (e.g., "Who is the author of the book?").
- Domain-specific Knowledge Inquiry: Questions requiring specialized background knowledge (e.g., "What style of architecture is this?").

STEP 3 - Populate the Example
Fill every key below using double quotes. Do not add extra keys.
{
 "description": "<STEP 1 output>",
 "task_type": "<Task selected in STEP 2>",
 "question": "<A visually grounded question in {language}>",
 "positive_answer": "<Concise, correct answer in {language}>",
 "hard_negative_answer": "<A plausible but deceptive incorrect answer in {language}>"
}

Hard Constraints:
- "task_type" must be exactly chosen from the list in STEP 2.
- Ensure the question is directly answerable from the visual or embedded textual content.
- Output ONLY the JSON object.
\end{verbatim}
\end{promptblock}

\begin{promptblock}{Video Classification Prompt}
\begin{verbatim}
You are given a video. Your job is to create ONE high-quality multimodal training example for a video classification dataset.  
The final answer MUST be a single JSON object and nothing else.

STEP 1 - Visual Analysis (less or equal than 300 {language} words)
- General overview of the video content.
- Identify primary actions, environmental settings, and the overall event type.
- Brainstorm potential ways this video could support the classification tasks listed in STEP 2.

STEP 2 - Task Selection
Choose ONE task type from the list below that best fits the video:
- Activity Recognition: Identifying the main activity or action being performed.
- Scene Parsing: Determining the primary environment or setting of the video.
- Event Categorization: Classifying the video into a specific event type or intended purpose.
- Sentiment/Intent Analysis: Recognizing the dominant emotional tone or the sentiment expressed.

STEP 3 - Populate the Example
Fill every key below using double quotes. Do not add extra keys.
{
  "description": "<STEP 1 output>",
  "task_type": "<Task Selected in STEP 2>",
  "label": "<Correct label in {language}>",
  "misleading_label": "<Plausible but incorrect label in {language} for hard negative mining>"
}

Hard Constraints:
- "task_type" must be exactly chosen from the list in STEP 2.
- "description", "label", and "misleading_label" must be in {language}.
- Output ONLY the JSON object—no extra text or explanations.
\end{verbatim}
\end{promptblock}

\section{Model Applications and Examples}

In this section, we present several real-world application scenarios to demonstrate the practical utility of Qwen3-VL-Embedding. The showcases in Table~\cref{tab:mm-examples-text,tab:mm-examples-image,tab:mm-examples-video,tab:mm-examples-visdoc} illustrate how the model handles diverse queries and complex visual data, providing a clearer understanding of its integration into downstream tasks.

\newcommand{\qimg}[1]{\includegraphics[width=0.45\linewidth]{#1}}
\newcommand{\dimg}[1]{\includegraphics[width=0.45\linewidth]{#1}}
\newcommand{\qvid}[1]{\includegraphics[width=0.95\linewidth]{#1}}
\newcommand{\dvid}[1]{\includegraphics[width=0.95\linewidth]{#1}}

\newcommand{\tagtxt}{\textbf{\small Text:}\ }
\newcommand{\tagimg}{\textbf{\small Image:}\ }
\newcommand{\tagvid}{\textbf{\small Video:}\ }

\newcolumntype{Y}{>{\raggedright\arraybackslash}X}

\newcommand{\cellt}[1]{%
  \begin{minipage}[t]{\linewidth}
    #1
  \end{minipage}%
}

\begin{table*}[htbp]
\centering
\small
\setlength{\tabcolsep}{6pt}
\renewcommand{\arraystretch}{1.2}

\caption{Similarity scores evaluated by Qwen3-VL-Embedding (text tasks).}
\label{tab:mm-examples-text}

\begin{tabularx}{\textwidth}{l Y}
\toprule
\rowcolor[HTML]{F5F5F5} \textbf{Task} & \textbf{AG News~\citep{zhang2015character}} \\
\textbf{Instruction} & Classify the news article. \\
\bottomrule
\end{tabularx}

\begin{tabularx}{\textwidth}{l X X c}
\toprule
\textbf{Ex.} & \textbf{Query} & \textbf{Document} & \textbf{Sim.} \\
\midrule

1 &
\cellt{%
  \tagtxt Fears for T N pension after talks Unions representing workers at Turner   Newall say they are 'disappointed' after talks with stricken parent firm Federal Mogul.
} &
\cellt{%
  \tagtxt Business
} &
0.55 \\

2 &
\cellt{%
  \tagtxt US fighter squadron to be deployed in South Korea next month (AFP) AFP - A squadron of US Air Force F-15E fighters based in Alaska will fly to South Korea next month for temporary deployment aimed at enhancing US firepower on the Korean peninsula...

} &
\cellt{%
  \tagtxt World
} &
0.57 \\

\bottomrule
\end{tabularx}

\begin{tabularx}{\textwidth}{l Y}
\toprule
\rowcolor[HTML]{F5F5F5} \textbf{Task} & \textbf{SQuAD~\citep{rajpurkar2016squad}} \\
\textbf{Instruction} & Retrieve passages that answer this question. \\
\bottomrule
\end{tabularx}

\begin{tabularx}{\textwidth}{l X X c}
\toprule
\textbf{Ex.} & \textbf{Query} & \textbf{Document} & \textbf{Sim.} \\
\midrule

1 &
\cellt{%
  \tagtxt Which NFL team represented the AFC at Super Bowl 50?
} &
\cellt{%
  \tagtxt Super Bowl 50 was an American football game to determine the champion of the National Football League (NFL) for the 2015 season. The American Football Conference (AFC) champion Denver Broncos defeated...
} &
0.81 \\

2 &
\cellt{%
  \tagtxt Who headlined the halftime show for Super Bowl 50?
} &
\cellt{%
  \tagtxt CBS broadcast Super Bowl 50 in the U.S., and charged an average of \$5 million for a 30-second commercial during the game. The Super Bowl 50 halftime show was headlined by the British rock group Coldpl...
} &
0.75 \\

\bottomrule
\end{tabularx}

\begin{tabularx}{\textwidth}{l Y}
\toprule
\rowcolor[HTML]{F5F5F5} \textbf{Task} & \textbf{MS MARCO~\citep{lin2014microsoft}} \\
\textbf{Instruction} & Retrieve relevant passages. \\
\bottomrule
\end{tabularx}

\begin{tabularx}{\textwidth}{l X X c}
\toprule
\textbf{Ex.} & \textbf{Query} & \textbf{Document} & \textbf{Sim.} \\
\midrule

1 &
\cellt{%
  \tagtxt walgreens store sales average
} &
\cellt{%
  \tagtxt The average Walgreens salary ranges from approximately \$15,000 per year for Customer Service Associate / Cashier to \$179,900 per year for District Manager. Average Walgreens hourly pay ranges from app...
} &
0.77 \\

2 &
\cellt{%
  \tagtxt how much do bartenders make
} &
\cellt{%
  \tagtxt According to the Bureau of Labor Statistics, the average hourly wage for a bartender is $10.36, and the average yearly take-home is $21,550. Bartending can be a lot of things. For some it is exciting,...
} &
0.81 \\

\bottomrule
\end{tabularx}

\end{table*}
\begin{table*}[htbp]
\centering
\small
\setlength{\tabcolsep}{6pt}
\renewcommand{\arraystretch}{1.2}

\caption{Similarity scores evaluated by Qwen3-VL-Embedding (image tasks).}
\label{tab:mm-examples-image}

\begin{tabularx}{\textwidth}{l Y}
\toprule
\rowcolor[HTML]{F5F5F5} \textbf{Task} & \textbf{CIFAR-10~\citep{krizhevsky2009learning}} \\
\textbf{Instruction} & Classify the object in this image. \\
\bottomrule
\end{tabularx}

\begin{tabularx}{\textwidth}{l X X c}
\toprule
\textbf{Ex.} & \textbf{Query} & \textbf{Document} & \textbf{Sim.} \\
\midrule

1 &
\cellt{%
  \tagimg \\
  \qimg{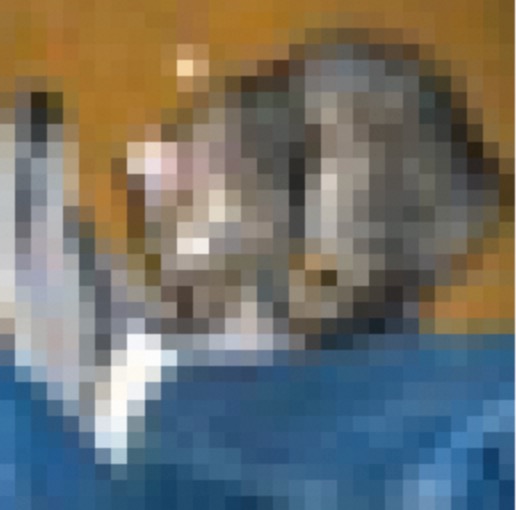}
} &
\cellt{%
  \tagtxt cat
} &
0.67 \\

2 &
\cellt{%
  \tagimg \\
  \qimg{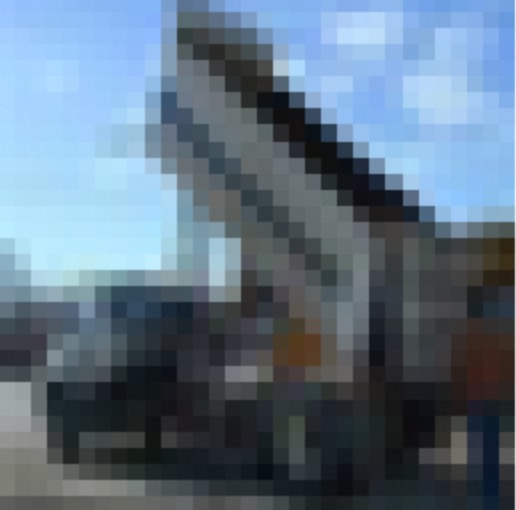}

} &
\cellt{%
  \tagtxt truck
} &
0.69 \\

\bottomrule
\end{tabularx}

\begin{tabularx}{\textwidth}{l Y}
\toprule
\rowcolor[HTML]{F5F5F5} \textbf{Task} & \textbf{VQAv2~\citep{goyal2017making}} \\
\textbf{Instruction} & Find the answer to this question about the image. \\
\bottomrule
\end{tabularx}

\begin{tabularx}{\textwidth}{l X X c}
\toprule
\textbf{Ex.} & \textbf{Query} & \textbf{Document} & \textbf{Sim.} \\
\midrule

1 &
\cellt{%
  \tagtxt Where is he looking?\\
  \tagimg\\
  \qimg{}
} &
\cellt{%
  \tagtxt down
} &
0.54 \\

2 &
\cellt{%
  \tagtxt What are the people in the background doing?\\
  \tagimg\\
  \qimg{figures/dataset_examples/vqa_v2/vqa_v2_0.jpg}
} &
\cellt{%
  \tagtxt watching
} &
0.67 \\

\bottomrule
\end{tabularx}

\begin{tabularx}{\textwidth}{l Y}
\toprule
\rowcolor[HTML]{F5F5F5} \textbf{Task} & \textbf{MS COCO~\citep{lin2014microsoft}} \\
\textbf{Instruction} & Find images matching this description. \\
\bottomrule
\end{tabularx}

\begin{tabularx}{\textwidth}{l X X c}
\toprule
\textbf{Ex.} & \textbf{Query} & \textbf{Document} & \textbf{Sim.} \\
\midrule

1 &
\cellt{%
  \tagtxt A man with a red helmet on a small moped on a dirt road.
} &
\cellt{%
  \tagimg\\
  \dimg{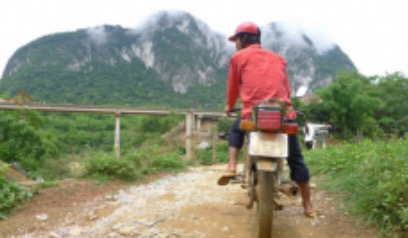}
} &
0.52 \\

2 &
\cellt{%
  \tagtxt The bathroom is clean and ready to be used.
} &
\cellt{%
  \tagimg\\
  \dimg{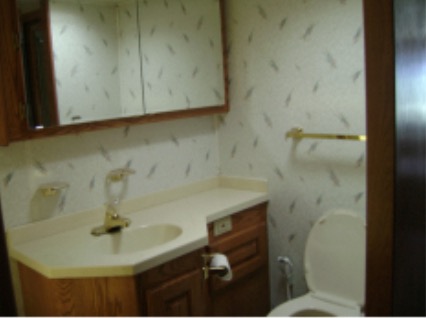}
} &
0.46\\

\bottomrule
\end{tabularx}

\end{table*}
\begin{table*}[htbp]
\centering
\small
\setlength{\tabcolsep}{6pt}
\renewcommand{\arraystretch}{1.2}

\caption{Similarity scores evaluated by Qwen3-VL-Embedding (video tasks).}
\label{tab:mm-examples-video}

\begin{tabularx}{\textwidth}{l Y}
\toprule
\rowcolor[HTML]{F5F5F5} \textbf{Task} & \textbf{UCF101~\citep{soomro2012ucf101}} \\
\textbf{Instruction} & Classify the action in this video. \\
\bottomrule
\end{tabularx}

\begin{tabularx}{\textwidth}{l X X c}
\toprule
\textbf{Ex.} & \textbf{Query} & \textbf{Document} & \textbf{Sim.} \\
\midrule

1 &
\cellt{%
  \tagvid \\
  \qvid{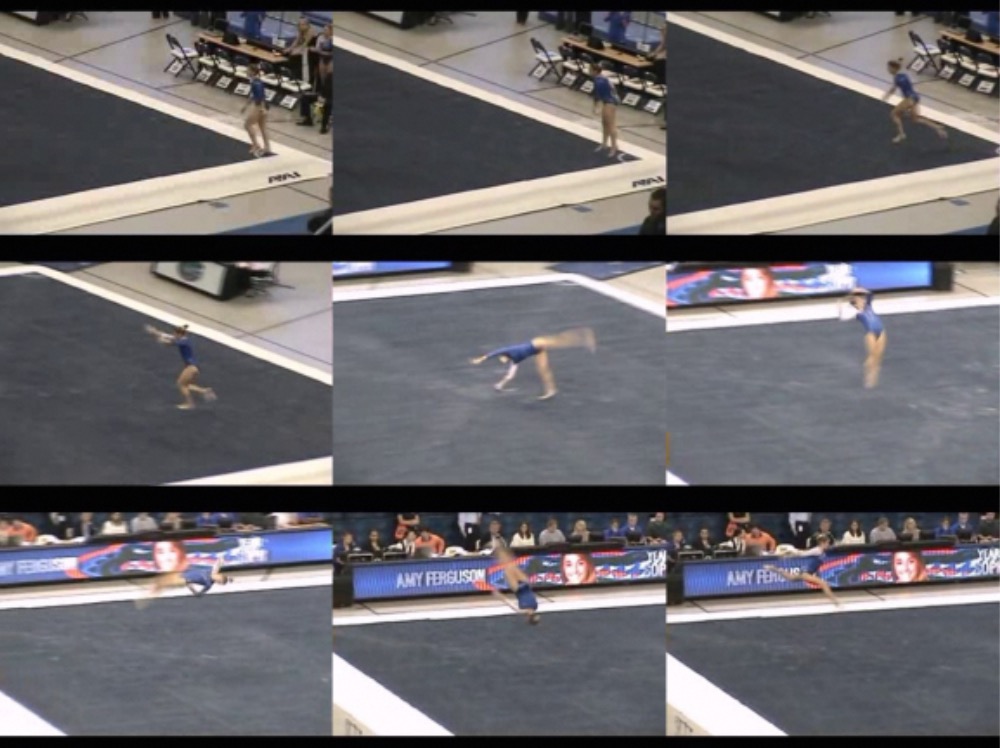}
} &
\cellt{%
  \tagtxt FloorGymnastics
} &
0.66 \\

\bottomrule
\end{tabularx}

\begin{tabularx}{\textwidth}{l Y}
\toprule
\rowcolor[HTML]{F5F5F5} \textbf{Task} & \textbf{NExTQA~\citep{xiao2021next}} \\
\textbf{Instruction} & Find the answer to this question about the video. \\
\bottomrule
\end{tabularx}

\begin{tabularx}{\textwidth}{l X X c}
\toprule
\textbf{Ex.} & \textbf{Query} & \textbf{Document} & \textbf{Sim.} \\
\midrule

1 &
\cellt{%
  \tagtxt Why did the girl have painted nail polish on her nails...\\
  \tagvid\\
  \qvid{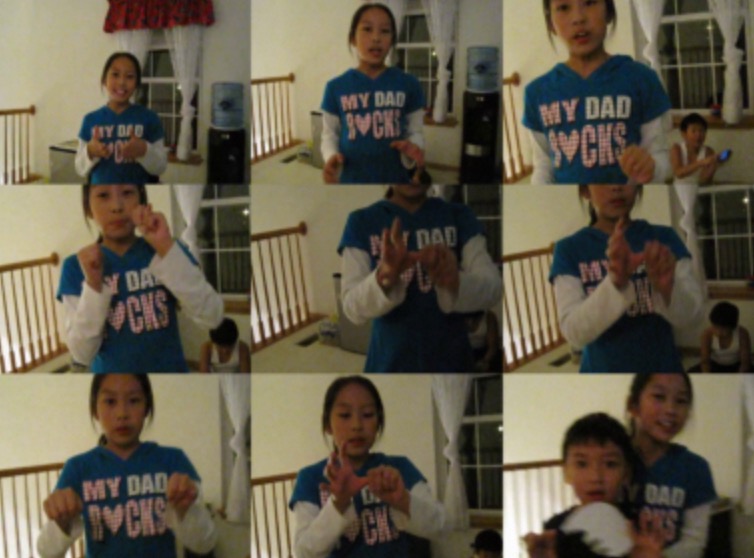}
} &
\cellt{%
  \tagtxt (E) look nice
} &
0.64 \\

\bottomrule
\end{tabularx}

\begin{tabularx}{\textwidth}{l Y}
\toprule
\rowcolor[HTML]{F5F5F5} \textbf{Task} & \textbf{MST-VTT~\citep{xu2016msr}} \\
\textbf{Instruction} & Find videos matching this description. \\
\bottomrule
\end{tabularx}

\begin{tabularx}{\textwidth}{l X X c}
\toprule
\textbf{Ex.} & \textbf{Query} & \textbf{Document} & \textbf{Sim.} \\
\midrule

1 &
\cellt{%
  \tagtxt baseball player hits ball
} &
\cellt{%
  \tagvid\\
  \dvid{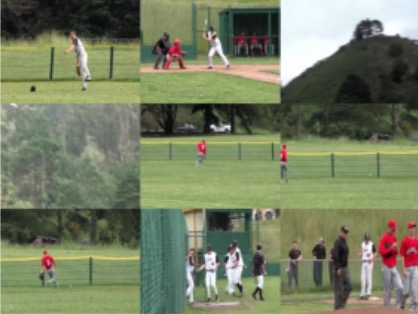}
} &
0.80 \\

\bottomrule
\end{tabularx}

\end{table*}
\begin{table*}[htbp]
\centering
\small
\setlength{\tabcolsep}{6pt}
\renewcommand{\arraystretch}{1.2}

\caption{Similarity scores evaluated by Qwen3-VL-Embedding (visual document tasks).}
\label{tab:mm-examples-visdoc}

\begin{tabularx}{\textwidth}{l Y}
\toprule
\rowcolor[HTML]{F5F5F5} \textbf{Task} & \textbf{ViDoRe\_ArxivQA~\citep{faysse2025colpali}} \\
\textbf{Instruction} & Find documents that answer this question. \\
\bottomrule
\end{tabularx}

\begin{tabularx}{\textwidth}{l X X c}
\toprule
\textbf{Ex.} & \textbf{Query} & \textbf{Document} & \textbf{Sim.} \\
\midrule

1 &
\cellt{%
  \tagtxt Based on the graph, what is the impact of correcting for fspec not equal to 1 on the surface density trend?
} &
\cellt{%
  \tagimg\\
  \dvid{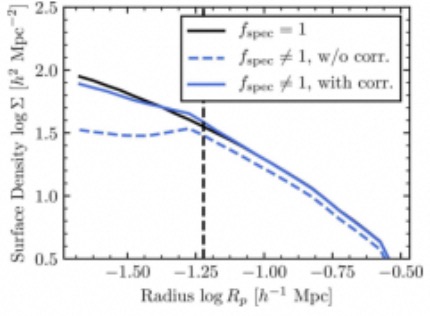}
} &
0.63 \\

2 &
\cellt{%
  \tagtxt Based on the progression from JUL10 to FEB11Q, what trend can be observed in the thread participation?
} &
\cellt{%
  \tagimg\\
  \dvid{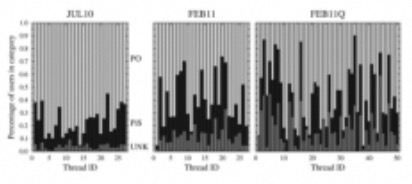}
} &
0.55 \\

\bottomrule
\end{tabularx}

\end{table*}

\end{document}